\documentclass{article}

\PassOptionsToPackage{numbers, compress}{natbib}
 \usepackage[preprint]{neurips_2026}

\usepackage[utf8]{inputenc} % allow utf-8 input
\usepackage[T1]{fontenc}    % use 8-bit T1 fonts
\usepackage{hyperref}       % hyperlinks
\usepackage{url}            % simple URL typesetting
\usepackage{booktabs}       % professional-quality tables
\usepackage{amsfonts}       % blackboard math symbols
\usepackage{nicefrac}       % compact symbols for 1/2, etc.
\usepackage{microtype}      % microtypography
\usepackage{xcolor}         % colors
\usepackage{amsmath}
\usepackage{booktabs}
\usepackage{pifont}
\usepackage{multirow}
\usepackage{graphicx}
\usepackage{xspace}
\usepackage{enumitem}
\usepackage{caption}
\usepackage{marvosym}

\newcommand{\cmark}{\ding{51}}
\newcommand{\xmark}{\ding{55}}
\newcommand{\ours}{{MAPLE}\xspace}

\makeatletter
\DeclareRobustCommand\onedot{\futurelet\@let@token\@onedot}
\def\@onedot{\ifx\@let@token.\else.\null\fi\xspace}
 
\def\ie{\emph{i.e}\onedot}

\title{MAPLE: Latent Multi-Agent Play for End-to-End Autonomous Driving}

\author{%
\textbf{Rajeev Yasarla$^{1\text{\Letter}}$\quad Deepti Hegde$^{1\star}$ \quad Hsin-Pai Cheng$^{1\star}$ \quad Shizhong Han$^{1\star}$ }\\
[1pt]
\textbf{Yunxiao Shi$^1$\quad Meysam Sadeghigooghari$^2$\quad Hanno Ackermann$^1$\quad Litian Liu$^1$}\\[1pt]
\textbf{Pranav Desai$^2$\quad Fatih Porikli$^1$\quad  Mohammad Ghavamzadeh$^1$\quad Hong Cai$^{1\text{\Letter}}$ }\\[5pt]
$^1$Qualcomm AI Research\thanks{Qualcomm AI Research is an initiative of Qualcomm Technologies, Inc.} \quad $^2$Qualcomm Technologies, Inc \\[3pt]
\small{\texttt{\Letter\{ryasarla,hongcai\}@qti.qualcomm.com}}\\[3pt]
\small{\textit{$^\star$Equal Contribution}}
}

\begin{document}

\maketitle

\begin{abstract}
Vision-language-action (VLA) models are effective as end-to-end motion planners, but can be brittle when evaluated in closed-loop settings due to being trained under traditional imitation learning framework. Existing closed-loop supervision approaches lack scalability and fail to completely model a reactive environment. We propose \ours, a novel framework for reactive, multi-agent rollout of a dynamic driving scenario in the latent space of the VLA model. The ego vehicle and nearby traffic agents are independently controlled over multi-step horizons, while being reactive to other agents in the scene, enabling closed-loop training. \ours consists of two training stages: (1) supervised fine‑tuning on the latent rollouts based on ground-truth trajectories, followed by (2) reinforcement learning with global and agent‑specific rewards that encourage safety, progress, and interaction realism. We further propose diversity rewards that encourage the model to generate planning behaviors that may not be present in logged driving data. Notably, our closed-loop training framework is scalable and does not require external simulators, which can be computationally expensive to run and have limited visual fidelity to the real-world. \ours achieves state-of-the-art driving performance on Bench2Drive and demonstrates scalable, closed-loop multi-agent play for robust E2E autonomous driving systems.
\end{abstract}

\section{Introduction}

% E2E, VLA
% End-to-end (E2E) motion planning has recently emerged as a new paradigm for autonomous driving. By optimizing perception, prediction, and planning under a unified framework, E2E models can potentially reduce error compounding and overcome limitations from hand-crafted interfaces. More recently, vision‑language‑action (VLA) models have shown great promises to significantly advance E2E autonomous driving. VLAs have the ability to ground language understanding in visual perception and action generation, and as such, can leverage the open-world knowledge and reasoning capabilities of powerful pretrained large-language models (LLMs) and vision-language models (VLMs). This makes it possible for VLAs to better handle rare, novel, and complex driving scenarios.  \comment{Since we do not actually use language prediction or grounding in our framework, maybe we should motivate the use of VLAs differently, like the LLMs are good for next token prediction and learning causal relationships between tokens.}

End-to-end (E2E) autonomous driving has emerged as a promising paradigm that unifies perception, prediction, and planning into a single learned model. Recent E2E planners, such as UniAD~\citep{hu2023planning} and VAD~\citep{jiang2023vad}, demonstrate strong performance under supervised learning, but often struggle to generalize to long-tail and interactive driving scenarios. Vision-language-action (VLA) models further advance this line of work by grounding driving behavior in language and leveraging the reasoning capabilities of large multimodal models. Approaches such as GPT-Driver~\citep{mao2023gpt}, EMMA~\citep{hwang2024emma}, DriveVLM~\citep{tian2024drivevlm}, and Senna~\citep{jiang2024senna} show improved robustness in complex driving situations, while ORION~\citep{fu2025orion} and SimLingo~\citep{renz2025simlingo} strengthen alignment between language and motion planning.

A central limitation of existing VLA planners is that training is primarily performed in an open-loop manner using supervised fine-tuning on large-scale logged driving data~\citep{fu2025orion,zhou2025autovla,renz2025simlingo}. As a result, these models do not explicitly model closed-loop interactions between the ego vehicle and other traffic participants, and typically treat surrounding agents as non-reactive during training. This gap becomes critical at deployment time, where driving is inherently closed-loop and multi-agent, requiring continuous adaptation to the evolving behavior of the other agents. Consequently, these models suffer from covariate shift, where small deviations from demonstrated trajectories compound over time and lead to planning failures~\citep{dauner2023parting,KarkusBeyondBC2025}. Additionally, ground-truth trajectory labels extracted from driving logs are very sparse, \ie, they only capture a very small number of possible trajectories in the action space, which makes it challenging to learn a robust policy from them alone~\citep{KarkusBeyondBC2025}. These limitations highlight the need for training mechanisms that explicitly support closed-loop, multi-agent interaction, ideally without relying on simulation pipelines that are costly and challenging to scale.

Reinforcement learning (RL) offers a natural way to perform closed-loop training by optimizing policies through interaction. Prior work has shown that large-scale RL can yield highly robust motion planners when trained with self-play and reactive agents in simulation~\citep{cusumanotowner2025robust,jaeger2025carl}. 
%\citep{silver2017mastering,vinyals2019grandmaster,openai2019dexterous,cusumanotowner2025robust,jaeger2025carl}.
However, these approaches rely on external simulators that operate on symbolic representations, such as maps, bounding boxes, and vehicle states~\citep{nuplan,waymax,suo2021trafficsim,bergamini2021itra,cusumanotowner2025robust}. While these symbolic simulators enable scalable RL, they are not directly applicable to end-to-end VLA models that operate on raw sensory inputs. Extending RL to E2E planners requires simulating reactive agents and future observations at the pixel level, which is computationally expensive and difficult to scale using graphics engines, splatting methods, or diffusion-based image generation~\citep{dosovitskiy2017carla,huang2025gen}. Moreover, pixel-level simulators often exhibit limited visual fidelity and distribution mismatch with real-world data, further hindering transfer. These challenges motivate our approach of enabling RL directly in the latent space of a VLA, avoiding pixel-level simulation while still supporting closed-loop, multi-agent interaction learning.

%Reinforcement learning is an effective tool in performing closed-loop training motion planners. Privileged motion planners such as GigaFlow and CARL are trained at scale to achieve highly generalized policies \cite{zulfiqar2025gigaflow,jaeger2025carl}. However, these planners are trained inside simulators that produce symbolic 2D map layouts of the scene, which are directly passed as input to the planners in the form of boxes, road boundaries, and vehicle states. The use of symbolic data allows the use of high-speed simulators that are ideal environments for scalable reinforcement learning and the modeling of reactive agents for closed-loop supervision. When applying this framework to end-to-end motion planners, a couple of challenges arise. Simulation of novel views and reactive agents at the image level is computationally expensive \cite{dosovitskiy2017carla} and has low visual fidelity to the real world \cite{dosovitskiy2017carla}\todo{[other neural rendering papers.]}. \comment{Need to update last sentence.}

Our goal is to enable this form of scalable closed-loop training for end-to-end motion planners. Towards this, we propose \ours, a framework that simulates the behavior of the ego-vehicle and multiple reactive agents in the latent space of a VLA model. This addresses the challenge of simulating sensory data while also training an E2E planner in a closed-loop environment in the presence of reactive agents. Diversity rewards encourage out-of-distribution behavior of surrounding agents at training time, providing richer feedback signals.

Our contributions are as follows.
\begin{itemize}
\item \textbf{Multi-agent scenario rollout in latent space across SFT and RL.} 
We introduce a simulator-free multi-agent rollout mechanism that jointly evolves the ego vehicle and surrounding reactive agents through autoregressive rollouts in the latent space of a VLA model. This enables closed-loop, multi-step agent interactions during training entirely in the latent space, significantly reducing computational cost compared to pixel-level training, while remaining removable at inference time without additional overhead.

\item \textbf{Diversity-aware RL for multi-agent play.}
To encourage heterogeneous and realistic interactions beyond logged driving data, we introduce a diversity-aware reward in the RL stage that explicitly promotes distinct agent behaviors. This leads to richer multi-agent interactions and more robust closed-loop driving policies.

\item \textbf{State-of-the-art performance on closed-loop autonomous driving benchmarks.}
Extensive experiments and ablation studies on the Bench2Drive benchmark show that the proposed multi-agent rollout mechanism and diversity-aware RL significantly improve driving performance, boosting the driving score by 25+\%, setting the new state-of-the-art, and enhancing robustness across complex multi-ability driving scenarios.
\end{itemize}

\section{Related Work}
\vspace{-10pt}
\paragraph{End-to-end and VLA models for autonomous driving.}
Early E2E planners such as UniAD \citep{hu2023planning} and VAD \citep{jiang2023vad} unify
perception, prediction, and planning in a single pipeline but struggle with long-tail
generalization. VLA models address this by grounding motion in language. GPT-Driver
\citep{mao2023gpt} casts planning as text generation with chain-of-thought reasoning
\citep{wei2022chain}. EMMA \citep{hwang2024emma} scales this to large multi-modal corpora.
DriveVLM \citep{tian2024drivevlm} and Senna \citep{jiang2024senna} pair a VLM with a
conventional planner or meta-action head. ORION \citep{fu2025orion} and SimLingo
\citep{renz2025simlingo} further tighten the language-trajectory alignment. 
ReCogDrive~\citep{li2026recogdrive} and DiffRefiner~\citep{yin2026diffrefiner} improve closed-loop robustness through recognition-driven planning and diffusion-based refinement, yet still rely on static logged trajectories for training. GeRo~\citep{yasarla2026generative} introduces language-conditioned latent rollouts
with ego planning and surrounding-agent motion prediction. Despite these
advancements, all these methods treat surrounding agents as
non-reactive, leaving closed-loop multi-agent dynamics underexplored. \ours closes this gap
by enabling reactive co-planning of ego vehicle and nearby agents entirely within a latent token space.

\paragraph{Multi-agent simulation and self-play.}
Trajectory forecasting methods \citep{helbing1995social, shi2022mtr, zhou2023qcnet} model
joint agent futures from fixed observations but do not support interactive planning. Self-play
has produced strong policies in games \citep{silver2017mastering, vinyals2019grandmaster} and
manipulation \citep{openai2019dexterous}. Recently, self-play has been applied to autonomous driving to obtain robust policies~\citep{cusumanotowner2025robust}. However, this is limited to planning in the symbolic space with bounding boxes, maps, and vehicle states. 
% yet its application in autonomous driving remains
% constrained by external simulators: TrafficSim \citep{suo2021trafficsim} and SimNet
% \citep{bergamini2021itra} simulate reactive agents in CARLA or nuPlan, while NAVSIM
% \citep{dauner2024navsim} provides non-reactive replay-based evaluation at scale. 
Adversarial scenario generation \citep{rempe2022generating, zhang2023cat} synthesizes
safety-critical behaviors but still relies on symbolic planners, not learned VLA policies. 
Extending self-play to E2E planners requires simulating future observations at the pixel level. Graphics engines~\citep{dosovitskiy2017carla}, Gaussian splatting, and diffusion-based generators~\citep{huang2025gen} can produce such inputs but suffer from high computational cost, limited visual fidelity, and poor scalability for interactive multi-agent rollouts. \ours provides a novel framework that enables multi-agent interactive play directly in the latent space of a VLA
model, requiring no external simulator or image rendering/generation. 
% \mgh{Should we call our method self-play? Moreover, make sure it is safe to say ``it is the first self-play ...''.}

\paragraph{Diverse action generation and RL-based policy optimization.}
Diffusion-based planners, including DiffusionDrive \citep{liao2025diffusiondrive}, DiffusionPlanner
\citep{zheng2025diffusion}, and GoalFlow \citep{xing2025goalflow}, capture multimodal
trajectory distributions through stochastic sampling but do not explicitly encourage
behavioral diversity during training, leaving them prone to mode collapse under RL
fine-tuning \citep{kirk2023survey}. GRPO \citep{shao2024deepseekmath} has been adapted for
driving in AlphaDrive \citep{jiang2025alphadrive}, R2SE \citep{liu2025reinforced}, and TrajHF
\citep{li2025finetuning}, improving policy stability via rule-based rewards, though each
assigns a single planner per agent. Sun et al.~\citep{sun2025latentcot} use latent token sequences and diversity-seeking rewards, focusing more on reasoning and not closed-loop driving. GenDrive \citep{huang2025gen} adds reward modeling to
a diffusion policy but still relies on a simulator. \ours addresses both gaps: it assigns
multiple discrete planners per agent and penalizes behavioral overlap with diversity-aware
GRPO rewards, preventing collapse and generating rich, safety-critical long-tail scenarios
without any external simulator.

% multi-agent rollout -> sft -> post training RL
% enable multi agent roll out in latent space.
% section 3: latent space multi-agent rollout.  seciton 4: training
\begin{figure*}[t]
    \centering
    \includegraphics[width=0.9\textwidth]{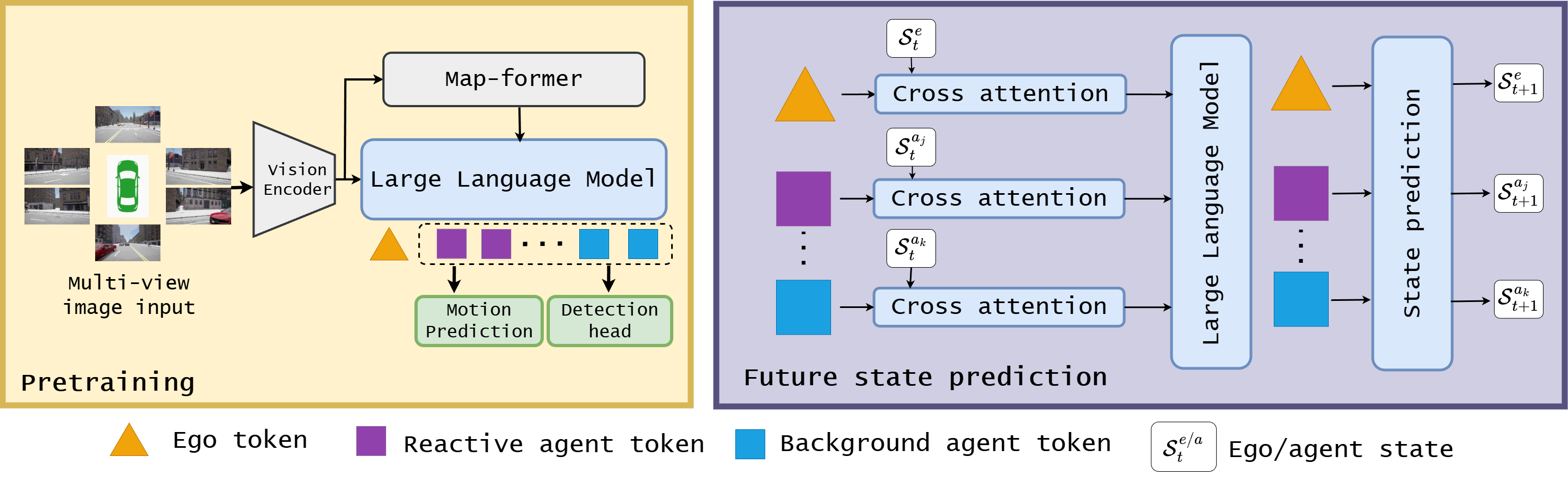}
    \vspace{-6pt}
    \caption{\small \textbf{\ours pretraining and future state prediction.}
    \textbf{Left:} Pretraining the VLA backbone with auxiliary supervision (e.g., map learning, detection, and motion prediction).
    \textbf{Right:} State-transition pretraining that predicts next-step ego/agent states over a horizon $T$ to stabilize the token space.} 
    \label{fig:ours_pretrain}
    \vspace{-10pt}
    % update S_hat t+1 at output
\end{figure*}

\section{Proposed Approach}\label{sec:method}

\begin{figure*}[t]
    \centering
    \includegraphics[width=0.9\textwidth]{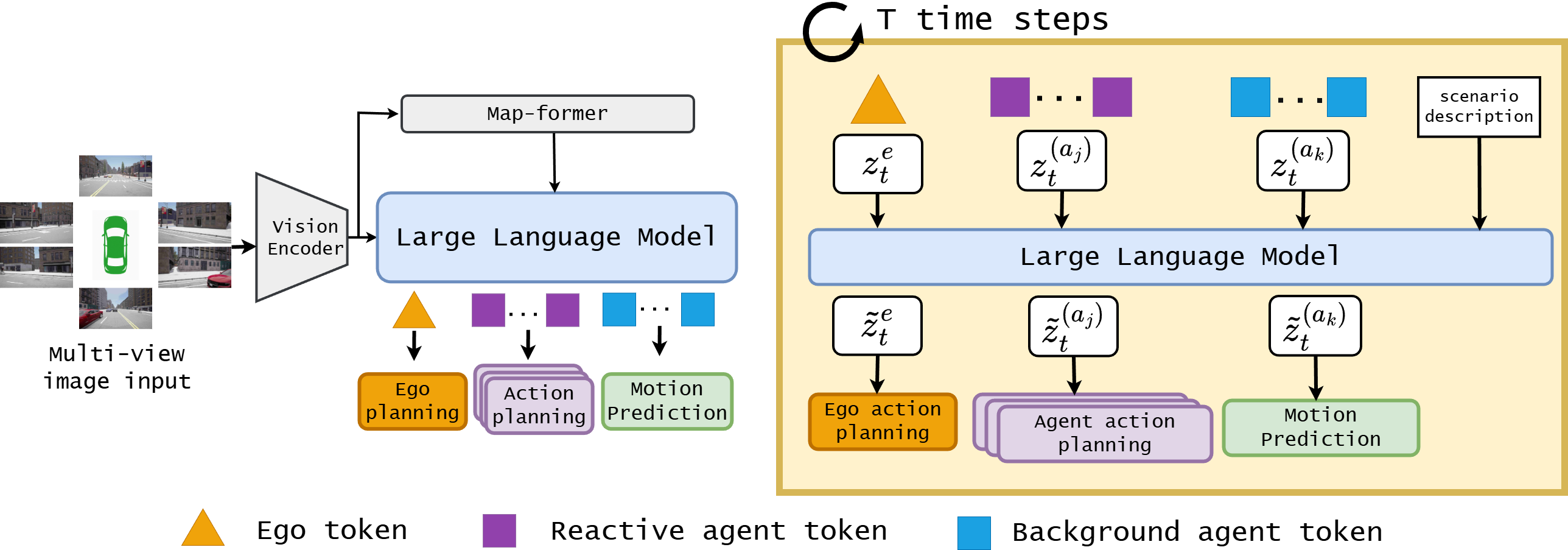}
    \vspace{-4pt}
    \caption{\small \textbf{\ours supervised fine-tuning (SFT) stage.}
    \textbf{Left:} Single-step supervision and inference. The VLA backbone encodes multi-view images (and map features) into ego and agent tokens, which are decoded by an ego planner, reactive-agent planners, and a motion head. 
    \textbf{Right:} The same model unrolled for $T$ steps during imitation-learning-based scenario rollouts. Predicted tokens/trajectories are fed back autoregressively to supervise ego and reactive-agent planning over the horizon while forecasting background-agent motion.}
    \label{fig:ours_sft}
    \vspace{-10pt}
    % fix notation
\end{figure*}

\begin{figure*}[t]
    \centering
    \includegraphics[width=0.6\textwidth]{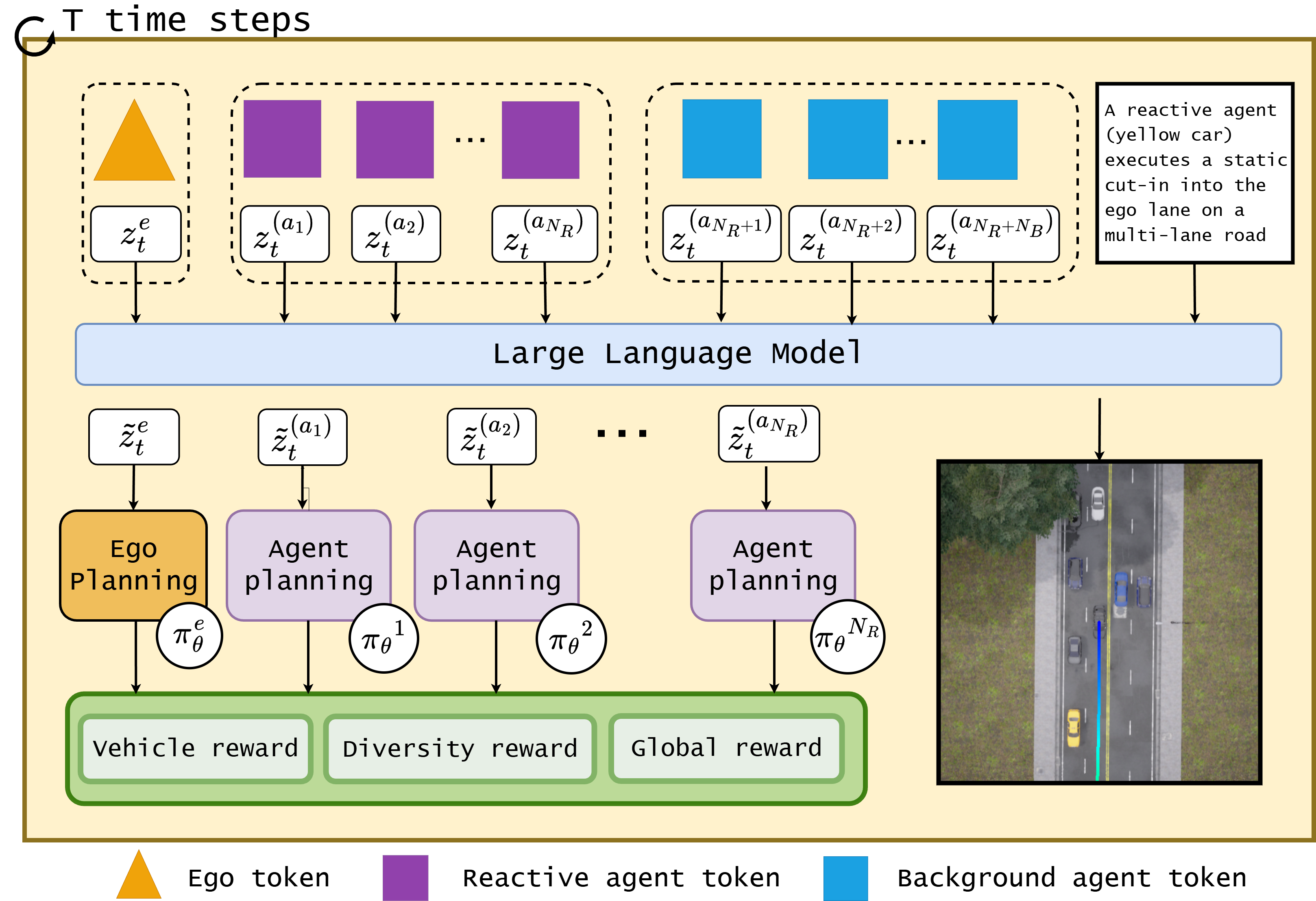}
    \vspace{-5pt}
    \caption{\small \textbf{\ours RL fine-tuning stage.}
    Starting from the SFT model, we optimize multi-step rollouts over $T$ steps using RL with safety-aware and interaction-aware rewards (e.g., collision avoidance and TTC).}
    \label{fig:ours_rl}
    \vspace{-10pt}
    % explain loss functions in words and not notations
\end{figure*}

We present \ours, a novel framework for end-to-end motion planning that performs closed-loop multi-agent rollout in the latent space of a VLA model. The ego vehicle and neighboring agents are represented as compact latent tokens, encoding velocity, acceleration, location, map labels, and traffic status. Given a latent ego vehicle token, latent agent tokens, and a high-level scenario description, \ours rolls out future scenarios by predicting future tokens in an autoregressive manner. This allows us to model the interaction of agents in the scene. Action planning and motion prediction heads are supervised with a series of rewards to encourage safe and diverse driving behavior. An overview of the multi-agent rollout mechanism in our proposed \ours framework can be seen in Figures~\ref{fig:ours_sft} and~\ref{fig:ours_rl}.

\subsection{Problem Setup}
Given multi-view camera images up to time $t$, our goal is to predict a safe ego-vehicle trajectory $\tau^e_{t:t+T}$, where $T$ is the planning horizon, while accounting for the future evolution of surrounding traffic participants. Unlike ego-only planners that model the behavior of the neighboring agents with ground-truth trajectory playback, we perform decision-making with \emph{reactive} agents, where the ego vehicle's plan is conditioned on the anticipated behavior of agents controlled by learned policies.

% (e.g., close proximity, high relative velocity, low TTC) --- move to appendix - criteria for assigning reactive vs background label

We divide the agents into (i) a set of \emph{reactive agents} ($\textbf{Z}^{\text{react}}_t$) that are likely to interact with the ego vehicle, and (ii) \emph{background agents} ($\textbf{Z}^{\text{bg}}_t$), whose behaviors are less likely to influence the ego vehicle. The future states of the ego and reactive agents are rolled out in an autoregressive manner, while the future trajectories of background agents are directly predicted through regression. 

% {commented this because we have not set up these terms yet and it is better to keep problem setup simple.} Concretely, we predict ego waypoints $\tau^e_{t:t+T}$ and reactive-agent trajectories $\{\tau^{(a_i)}_{t:t+T}\}_{a_i \in \textbf{Z}^{\text{react}}_t}$, and we forecast motion trajectories $\{\nu^{(a_j)}_{t:t+T}\}_{a_j \in \textbf{Z}^{\text{bg}}_t}$ for the remaining agents.

\subsection{\ours Architecture}
\vspace{-4pt}
% \ours is a plug-and-play training framework for Vision Language Action (VLA) models.
% The vision encoder and text tokenizer map multi-view images and text prompts into a shared embedding space. A central LLM then fuses these multimodal tokens to produce compact latent representations for the ego vehicle and surrounding agents, enabling temporally consistent token-space rollouts. In addition, the LLM generates a scene-level description and can answer behavior-centric questions to provide language grounding during training.
% Finally, the framework includes (i) generative action-planning heads (implemented as a VAE) that decode ego and reactive-agent tokens into corresponding waypoint trajectories, and (ii) a motion-prediction head that decodes agent tokens into multi-agent trajectories.

\ours is a closed-loop training framework for VLA models. The framework consists of a vision encoder, a map encoder, an LLM, and several task-specific heads related to object detection, map segmentation, motion prediction, and action planning. The action planning head is a variational auto-encoder (VAE) that decodes ego and reactive agent tokens into their respective trajectory waypoints~\citep{fu2025orion}. The motion prediction head is a multi-layer perceptron that regresses future trajectories from background agent tokens. An overview of the architecture can be seen in Figure \ref{fig:ours_sft}.

\subsection{Pretraining}
In order to learn a stable tokenization of ego and agent dynamics for consistent autoregressive scenario generation, the VLA model is first pretrained to perform  the auxiliary tasks of 3D object detection, map segmentation, agent motion prediction, and traffic light segmentation. Detailed definitions of all auxiliary pretraining losses and their weighting are provided in Appendix~\ref{app:losses_pretrain}. These tasks encourage the shared backbone to encode geometry, semantics, and traffic context into latent agent tokens that can be reliably decoded over time. The model is also trained to predict the future states of the ego vehicle and the surrounding agents. An overview of the \ours future state prediction is shown in Figure~\ref{fig:ours_pretrain}.

\paragraph{Future state prediction.}
The state of a traffic participant at time $t$ is represented by four ordered tokens: $S_t = \Big\langle \texttt{<DYN>},\ \texttt{<TYPE>},\ \texttt{<MS>},\ \texttt{<TS>} \Big\rangle_t,$
where \texttt{<DYN>} encodes agent dynamics (e.g., position, heading, velocity, acceleration), \texttt{<TYPE>} is a categorical token in
$\{\texttt{car}, \texttt{truck}, \texttt{pedestrian}, \texttt{cyclist}\}$, \texttt{<MS>} is the index of the map segment where the agent resides, and \texttt{<TS>} encodes traffic light or traffic status information related to the agent and associated to the selected map context.
Given the current state $S_t$ (and the corresponding ego/agent token from  VLA encoder), the model predicts the next state $\hat{S}_{t+1}$.
% By recursively applying this transition, we perform future state-prediction for $T$ steps to obtain $\hat{S}^{(a_i)}_{t:t+T}$ and supervise with ground-truth states $S^{(a_i)}_{t:t+T}$.

We define the future state prediction loss $\mathcal{L}_{\text{state}}$ as a weighted combination of
(i) cross-entropy losses for the discrete state labels $\texttt{<MS>}$ and $\texttt{<TS>}$,
and (ii) an $\ell_1$ regression loss for the continuous dynamics token $\texttt{<DYN>}$:
\setlength{\abovedisplayskip}{2pt}
\setlength{\belowdisplayskip}{2pt}
\begin{equation}
\label{eq:fse_loss}
\mathcal{L}_{\text{state}}
= \lambda_{\text{dyn}}\big\|\widehat{\texttt{<DYN>}} - \texttt{<DYN>}\big\|_1
+ \lambda_{\text{ms}}\,\mathrm{CE}\!\big(\widehat{\texttt{<MS>}}, \texttt{<MS>}\big)
+ \lambda_{\text{ts}}\,\mathrm{CE}\!\big(\widehat{\texttt{<TS>}}, \texttt{<TS>}\big),
\end{equation}
where $\widehat{\cdot}$ and $\mathrm{CE}(\cdot,\cdot)$ denote model predictions and cross-entropy loss, and $\lambda_{\text{dyn}}$, $\lambda_{\text{ms}}$, and $\lambda_{\text{ts}}$ are the weights for the loss terms. The overall pretraining objective consists of minimizing the sum of the respective losses on 3D object detection, map segmentation, motion prediction, and future state prediction.
\vspace{-6pt}
% \paragraph{Pretraining objective.}
% The overall pretraining objective combines detection loss $\mathcal{L}_{\text{det}}$, motion loss $\mathcal{L}_{\text{mot}}$, map segmentation loss $\mathcal{L}_{\text{map}}$, and state-transition loss $\mathcal{L}_{\text{state}}$:
% \begin{equation}\label{eqn:pretraining}
% \mathcal{L}_{\text{pre}}
% = \mathcal{L}_{\text{det}} + \mathcal{L}_{\text{mot}} + \mathcal{L}_{\text{map}} + \mathcal{L}_{\text{state}}.
% \end{equation}
% This initialization yields temporally consistent agent representations and stabilizes downstream autoregressive scenario rollouts and GRPO-based reinforcement fine-tuning.

\subsection{\ours Rollout and Training Framework}
The closed-loop framework enables language-conditioned scenario rollouts for the ego vehicle and a subset of \emph{reactive} agents. The objective is to learn a compact token space in which the VLA backbone can roll out future traffic states in an autoregressive manner and support behavior-aware action planning over a horizon $T$. Compared to ego-only planners, \ours explicitly models multi-step interactions by (i) jointly rolling out ego and reactive-agent states, and (ii) encouraging diverse driving behaviors via safety and diversity rewards.

\subsubsection{Multi-Agent Rollout and Supervised Fine-Tuning}
Given a set of historical and current multi-view images at time $t$ and a high-level scenario description, \ours encodes the scene into a latent ego token $z_t^e$ , a set of latent reactive agent tokens $\textbf{Z}^{\text{react}}_t$ indexed by $a_j$, and a set of latent background agent tokens $\textbf{Z}^{\text{bg}}_t$ indexed by $a_k$ (see Figure~\ref{fig:ours_sft}). Auto-regressive rollout is performed for $T$ steps. At each rollout step $\Delta = \{0,\ldots,T-1\}$, the VLA model predicts the next-step tokens $\tilde{z}_{t+\Delta+1}$ conditioned on the current tokens $\tilde{z}_{t+\Delta}$ and the scenario description. Here $\tilde{z}_t$ denotes latent tokens generated by auto-regressive rollout, distinguished from ground-truth tokens $z_t$ obtained from logged data.

The predicted tokens are decoded into trajectories and fed back as input for the next rollout step, yielding a rollout from $t$ to $t+T$. The planning heads produce waypoint trajectories for the ego vehicle and reactive agents:
the ego planner $P^e$ decodes $\tilde{z}_{t+\Delta+1}^e$ into trajectory $\tilde{\tau}^e_{t+\Delta+1}$,
and agent planners $\{P^{(a_j)}\}_{a_j \in \textbf{Z}^{\text{react}}_t}$ decode reactive-agent tokens into trajectories $\{\tilde{\tau}^{(a_j)}_{t+\Delta+1}\}$.
Let
$
\mathcal{P} \;=\; \{P^e\} \cup \{P^j\}_{j=1}^{N_R}
$
denote the set of action planners, where $N_R$ denotes the number of reactive-agent planners. At the start of each rollout, we use $P^e$ as the ego action planner, and for each reactive agent we select an action planner based on the agent's behavior.

The motion head predicts future trajectories for background agents indexed by $a_k$. Background agents participate in the latent rollout, however, they are \emph{not controlled} by planning heads. Instead, their trajectories $\{\tilde{\tau}^{(a_k)}_{t+\Delta+1}\}$ are generated by a motion prediction head conditioned on the current scene tokens. This design models passive scene dynamics while reserving explicit decision-making for the ego vehicle and reactive agents.

% The predicted tokens are then fed back as input for the next rollout step, yielding a rollout from $t$ to $t+T$.

\paragraph{Supervised Fine-Tuning (SFT).} In the SFT stage, we perform imitation-learning-based scenario rollouts for $T$ steps, supervising the ego vehicle and a set of reactive agents, while predicting motion for background agents. We supervise each step using ground-truth trajectories, providing accurate targets for the rollout horizon. The SFT objective loss is defined as the sum of per-step planning and motion prediction losses over the rollout horizon.
\setlength{\abovedisplayskip}{2pt}
\setlength{\belowdisplayskip}{2pt}
\begin{equation}
\label{eq:sft_loss}
\mathcal{L}_{\text{SFT}}
=
\sum_{\Delta=1}^{T}
\Bigg[
\mathcal{L}_{\text{plan}}\!\big(\tilde{\tau}^e_{t+\Delta}, \tau^e_{t+\Delta}\big)
+
\sum_{a_j \in \textbf{Z}^{\text{react}}_t}
\mathcal{L}_{\text{plan}}\!\big(\tilde{\tau}^{(a_j)}_{t+\Delta}, \tau^{(a_j)}_{t+\Delta}\big)
+
\sum_{a_k \in \textbf{Z}^{\text{bg}}_t}
\mathcal{L}_{\text{mot}}\!\big(\tilde{\tau}^{(a_k)}_{t+\Delta}, \tau^{(a_k)}_{t+\Delta}\big)
\Bigg].
\end{equation}  

% \comment{Is it necessary to notate trajectories $\tau$ here? We differentiate by $a_j$ and $a_k$ as well as $\mathcal{L}_{plan}$ and $\mathcal{L}_{mot}$, we can just say $\mathcal{L}^{ego}_{plan}(t+\Delta)$,$\mathcal{L}^{react}_{plan}$,  and $\mathcal{L}^{bg}_{mot}$.}

Here, by $\mathcal{L}_{\text{plan}}(\cdot,\cdot)$, we denote the planning loss used for supervising ego and reactive-agent trajectories, and by $\mathcal{L}_{\text{mot}}(\cdot,\cdot)$, the motion prediction loss applied to background agents. The detailed formulation of the planning loss is provided in Appendix~\ref{app:losses_planner}.

\subsubsection{Post-Training Reinforcement Learning}\label{sec:rl_posttrain}
The main objective of \ours is to learn \emph{diverse} and \emph{robust} driving behaviors while performing multi-step scenario rollouts that induce a wide range of environmental dynamics and ego--agent interactions.
In such settings, expert demonstrations are often multi-modal (e.g., brake vs.\ yield vs.\ merge), and pure imitation learning can be sub-optimal or even conflictual under distribution shift.
Thus, relying solely on SFT (Eq.~\ref{eq:sft_loss}) is insufficient to handle long-tail events and novel reactive interactions. To mitigate these issues and encourage safe, high-fidelity, and behaviorally diverse rollouts, we introduce an RL fine-tuning stage using the Group Relative Policy Optimization (GRPO) algorithm~\citep{shao2024deepseekmath} with a structured reward design.

\paragraph{Reward structure.}
At each rollout step $t+\Delta$, we compute three types of rewards:
(i) a \textbf{global reward} that captures scene-level safety and stability,
(ii) a \textbf{vehicle-specific reward} that encourages progress and safe driving for each controlled agent, and
(iii) a \textbf{diversity reward} that promotes distinct behaviors across different planners/policies.

% Let $\mathcal{I}_t = \{e\} \cup \textbf{Z}^{\text{react}}_t$ denote the set of \emph{controlled} agents (ego + reactive agents).
At a time $t$, we define the total rollout reward as
\setlength{\abovedisplayskip}{2pt}
\setlength{\belowdisplayskip}{2pt}
\begin{equation}
\label{eq:total_return}
R_t
=
G_t
+
\,D_t
+
r^{e}_t
+
\sum_{a_j \in \textbf{Z}^{\text{react}}} r^{a_j}_t,
\end{equation}
where $G_t$ and $D_t$ are global and diversity rewards, $r^e_t$ is the vehicle-specific reward for the ego vehicle, and $r^{a_j}_t$ is the vehicle-specific reward for the reactive agent $a_j \in \textbf{Z}^{\text{react}}$. %, and $D_t$ is the diversity reward. 
\paragraph{\textbf{Global reward.}}
The global reward encourages long collision-free rollouts and penalizes scene-level infractions.
Let $l$ be the number of steps completed without collision (early termination yields smaller $l$) in a $T$-step rollout.
We define the global reward $G_t$ at time $t$ as
\setlength{\abovedisplayskip}{2pt}
\setlength{\belowdisplayskip}{2pt}
\begin{equation}
\label{eq:global_reward}
G_t
=
\frac{l}{T}
-
\sum_{\Delta=1}^{T} \mathrm{Col}\!\left(\tilde{\tau}_{t+\Delta}^{e}\right)
-
\sum_{a_j \in \textbf{Z}^{\text{react}}}
\sum_{\Delta=1}^{T}
\mathrm{Col}\!\left(\tilde{\tau}_{t+\Delta}^{a_j}\right),
\end{equation}
where $\mathrm{Col}(\cdot)$ is the collision penalty calculated from the predicted trajectory $\tilde{\tau}_{t+\Delta}$. 

\paragraph{\textbf{Vehicle-specific reward.}}
Vehicle-specific rewards promote route completion/progress while discouraging unsafe behaviors, such as low time-to-collision (TTC). We use a step-wise reward and sum it over the rollout horizon $T$:
\setlength{\abovedisplayskip}{2pt}
\setlength{\belowdisplayskip}{2pt}
\begin{equation}
\label{eq:agent_reward}
r_t
=
\sum_{\Delta=1}^{T}
\Big[
\underbrace{RC_{t+\Delta}}_{\text{route completion}}
-
\,\underbrace{\phi\!\left(\mathrm{TTC}_{t+\Delta}\right)}_{\text{TTC penalty}}
-
\,\underbrace{\mathcal{L}_{\text{prog}}\!\left(\tilde{\tau}_{t+\Delta}\right)}_{\text{progress loss}}
\Big].
\end{equation}
%where $RC_{t+\Delta}^i$ measures incremental route completion, $\mathcal{L}_{\text{prog}}$ measures deviation from a desired route/baseline (or lack of forward progress), and $\phi(\mathrm{TTC})$ is a monotone penalty ( $\phi(\mathrm{TTC}) =  \max(0, \mathrm{TTC}_{\max}-\mathrm{TTC})$).
%
Here $RC_{t+\Delta}$ measures incremental progress along the reference route,
$\mathcal{L}_{\text{prog}}$ penalizes lack of forward progress or deviation from a desired trajectory,
and $\phi(\mathrm{TTC})$ is a monotone penalty function that is activated when TTC falls below a safety threshold, e.g.,~$\phi(\mathrm{TTC}) = \max(0, \mathrm{TTC}_{\max}-\mathrm{TTC})$.

\paragraph{\textbf{Diversity reward.}}
Safety and progress objectives can favor similar policies, so multiple planning heads alone do not ensure heterogeneous agent responses. The diversity reward $D_t$ discourages behavioral collapse by rewarding separation in driving-quality descriptors. This exposes the ego policy to a wider range of nearby-agent responses during rollout, helping it to learn decisions that accommodate different yielding, compliance, and comfort behaviors. Global collision and vehicle-specific TTC/progress rewards provide complementary safety and task feedback.
% To encourage diverse driving behaviors (e.g., conservative versus assertive merges) and prevent policy collapse, we introduce a diversity reward over trajectories induced by different action planners during rollout. 

We characterize the behavior of the ego and each of the $N_R$ reactive agents $a_j \in \textbf{Z}^{\text{react}}$ using a compact descriptor formed from NAVSIM driving-quality metrics~\cite{dauner2024navsim}, which captures salient properties, such as mean acceleration, jerk, minimum time-to-collision (TTC), lane centering error, and lane-change timing. We denote these compact descriptors by $\Gamma^e(\cdot)$, for the ego, and by $\{\Gamma^{a_j}(\cdot)\}_{j=1}^{N_R}$, for the reactive agents. Note that if two agents belong to the same behavior category, then they will have the same compact descriptor.   

%We summarize each planner-induced trajectory using a compact behavior descriptor $\Gamma(\cdot)$ formed from NAVSIM driving-quality metrics~\cite{dauner2024navsim}, which captures salient properties, such as mean acceleration, jerk, minimum time-to-collision (TTC), lane centering error, and lane-change timing.

At the start of the rollouts, the ego vehicle is controlled by a policy/planner $\pi_\theta^e$, and the reactive agents are controlled by policies $\{\pi_\theta^{a_j}\}_{j=1}^{N_R}$. For each of these agents, we generate a rollout/trajectory over the horizon $T$ using its own policy, i.e., $\{\tilde{\tau}^{a_j}_{t+\Delta}\}_{\Delta=1}^T \sim \pi_\theta^{a_j}$. The diversity reward for the ego (and similarly for each reactive agent) is computed as   
%
%, denote the planner index selected for $N_R$ reactive agents. For each reactive agent $a_j \in \textbf{Z}^{\text{react}}_t$, we select a planner $P^{\pi^{(a_j)}}$  conditioned on the agent’s behavior category and generate a corresponding trajectory over a horizon $T$.
%The resulting behavior descriptors are then used to optimize reactive-agent policies to increase behavioral diversity across planner-induced rollouts.
%Let $\mathcal{P} = P^e \cup P^{\pi(a_j)}$. We define the diversity reward as the average pairwise $\ell_1$ distance between behavior descriptors across different behavior metrics:
% \begin{equation}
% \label{eq:div_reward}
% D_t
% =
% \frac{1}{N_R(N_R-1)}
% \sum_{k \neq k'}
% \left\|
% \Gamma\!\left(\tilde{\tau}^{(a_k)}_{t:t+T}\right)
% -
% \Gamma\!\left(\tilde{\tau}^{(a_{k'})}_{t:t+T}\right)
% \right\|_1,
% \end{equation}
\setlength{\abovedisplayskip}{2pt}
\setlength{\belowdisplayskip}{2pt}
\begin{equation}
\label{eq:div_reward}
D_t
=
\frac{1}{N_R}
\sum_{j=1}^{N_R} \sum_{\Delta=1}^T
\Big\|\Gamma^e\big(\tilde{\tau}^e_{t+\Delta}\big)
-
\Gamma^{a_j}\big(\tilde{\tau}^{a_j}_{t+\Delta}\big)
\Big\|_1,
\end{equation}
which assigns higher rewards when different planner selections induce measurably distinct (yet safe) behaviors.
Additional details on the descriptor $\Gamma(\cdot)$, behavior category, and the diversity reward are provided in Appendix~\ref{app:diversity_reward}.

\paragraph{GRPO objective.}
Given a group of $G$ rollouts at time $t$ with returns $\{R_t^{(g)}\}_{g=1}^{G}$, we compute a group-relative baseline
$b_t = \frac{1}{G}\sum_{g=1}^{G} R_t^{(g)}$ and advantages $A_t^{(g)} = R_t^{(g)} - b_t$.
We then optimize the policy using the GRPO loss:
\begin{equation}
\label{eq:grpo_loss}
\mathcal{L}_{\text{RL}}
=
- \frac{1}{G}\sum_{g=1}^{G}\min\Big(r_\theta^{(g)}A_t^{(g)},\;
\operatorname{clip}\big(r_\theta^{(g)},1-\epsilon,1+\epsilon\big)A_t^{(g)}\Big),
%A_t^{(g)}\sum_{\Delta=1}^{T}\log \pi_\theta\!\left(x^{(g)}_{t+\Delta}\mid \tilde{z}^{(g)}_{t+\Delta}, s_t\right),
\end{equation}
where $r_\theta^{(g)}=\frac{\pi_\theta\big(\{\tilde{\tau}^{(g)}_{t+\Delta}\}\big)}{\pi_{\theta_{\text{old}}}\big(\{\tilde{\tau}^{(g)}_{t+\Delta}\}\big)}$ and the clip parameter is set to \(\epsilon=0.1\).

\vspace{-0pt}
\section{Experimental Setup}\label{sec:exp_setup}
% \subsection{Datasets and Evaluation Metrics}
\vspace{-7pt}
\paragraph{Datasets.}
We conduct both training and evaluation of \ours using the Bench2Drive benchmark~\citep{jia2024bench2drive}, a closed-loop end-to-end autonomous driving suite built on the CARLA simulator~\citep{dosovitskiy2017carla}. Following the official benchmark setup, the dataset consists of 1,000 driving clips, among which 950 are used for training and 50 are reserved for open-loop validation. Each clip spans roughly 150 meters and captures a wide range of traffic conditions and interaction patterns. Closed-loop performance is assessed using the standard Bench2Drive protocol, which evaluates agents over 220 routes covering 44 diverse interactive scenarios. We additionally evaluate on the in-distribution and generalization splits of Fail2Drive~\citep{Gerstenecker2026Fail2Drive} for closed-loop driving under distribution shifts.
% \vspace{-8pt}
\paragraph{Evaluation Metrics.}
On Bench2Drive, we adopt the official closed-loop evaluation criteria, including Driving Score (DS), Success Rate (SR), Efficiency, Comfort, and Multi‑Ability~\citep{jia2024bench2drive}. Driving Score reflects overall route completion while accounting for traffic violations. Success Rate reports the percentage of routes finished without failure. Efficiency and Comfort quantify the agent’s driving speed and smoothness, respectively. The Multi‑Ability metric further measures performance across five complex urban driving behaviors. 
\paragraph{Model Architecture.}
We instantiate \ours on the multimodal LLM backbone Qwen2.5-1.5B-VL~\citep{bai2023qwenvl} for end-to-end multi-agent planning. The Qwen2.5 text encoder and LLM generate scenario-level text tokens and ego/agent-specific latent tokens. For visual inputs, \ours employs an EVA-pretrained vision transformer (ViT)~\citep{fang2023eva} with a Q-Former module~\citep{fu2025orion} to extract image tokens from multi-view observations. The detection and motion prediction heads each consist of a multi-head attention module followed by three MLP layers. The future state prediction head adopts the same architecture. Each action-planning head in the planner set $\mathcal{P}$ is implemented as a variational autoencoder (VAE), following existing VLA planning heads~\citep{fu2025orion}. Unless stated otherwise, \ours predicts trajectories conditioned only on Navigation Command (NC), without explicit lane-center targets, such as Target Point (TP). \ours adopts an anchor-free design and predicts six trajectory modes aligned with Bench2Drive. Scenario rollouts are used \emph{only during training} and are disabled at inference time.

% \vspace{-8pt}
\paragraph{Training.}
All experiments are conducted on 8 NVIDIA H100 GPUs. \ours is trained in three stages. (1) Pretraining: 18 epochs with auxiliary supervision, including detection, map segmentation, motion prediction, and future state prediction, to learn robust latent token representations. (2) Supervised Fine-Tuning (SFT): 24 epochs of scenario rollouts with imitation learning. Autoregressive rollouts take $T$ steps, supervising ego-vehicle and reactive-agent planning and background-agent motion prediction. (3) Post-Training Reinforcement Learning: 6 epochs using the proposed reward mechanism (Section~\ref{sec:rl_posttrain}), with $T$-step autoregressive rollouts optimizing ego and reactive-agent planning via agent-specific, global, and diversity rewards. We use 4 input frames, set $N_R=8$ for reactive-agent planners, and use a rollout horizon of $T=8$. To increase temporal coverage, we sample rollouts with multiple temporal strides corresponding to effective rates of 0.5Hz, 1Hz, 1.5Hz, 2Hz, 5Hz, and 10Hz, covering up to 12 seconds into the future. All Qwen2.5‑1.5B parameters are updated across stages. Optimization uses AdamW with cosine annealing, a learning rate of $2\times10^{-4}$, and weight decay 0.01.
\begin{table*}[t!]
\centering
\caption{\small Comparison of closed-loop planning and multi-ability performance on the Bench2Drive. Avg.~L2 denotes the average trajectory error over 2 seconds at 2~Hz. Here Cond. is the applied condition of  NC = Navigation Command or TP = Target Point. $\dagger$ indicates models trained with geometric path waypoint supervision and evaluated using the PID controller from SimLingo.}
\label{tab:b2d_unified}
\vspace{-5pt}
\resizebox{0.98\textwidth}{!}{
\begin{tabular}{lp{0.9cm}cccccccccccc}
\toprule
\multirow{2}{*}{Method} & \multirow{2}{*}{Cond.} & 

\multicolumn{4}{c}{Closed-Loop Metrics} &
\multicolumn{6}{c}{Multi-Ability Test (\%) $\uparrow$} \\
\cmidrule(lr){3-6}
\cmidrule(lr){7-12}
 & &
DS$\uparrow$ &
SR (\%)$\uparrow$ &
Efficiency$\uparrow$ &
Comfort$\uparrow$ &

Merging &
Overtaking &
E-Brake &
Give Way &
T.Sign &
Mean \\

\midrule
TCP-traj~\cite{wu2022trajectory} & TP &

59.9 & 30.0 & 76.5 & 18.1 & 
28.8 & 24.3 & 51.7 & 40.0 & 46.3 & 34.2 \\

% TCP-traj w/o distillation~\cite{wu2022trajectory} &

% 49.3 & 20.5 & 78.8 & 22.9 &
% 28.7 & 28.7 & 48.3 & 40.0 & 28.7 & 34.1 \\

ThinkTwice~\cite{jia2023thinktwice} & TP &

62.4 & 31.2 & 69.3 & 16.2 &
27.4 & 18.4 & 35.8 & 50.0 & 54.2 & 37.2 \\

DriveAdapter~\cite{jia2023driveadapter} & TP &

64.2 & 33.1 & 70.2 & 16.0 & 
28.4 & 28.4 & 47.5 & 50.0 & 56.4 & 42.1 \\

SimLingo$\dagger$~\cite{renz2025simlingo} & TP &

85.1 & 67.3 & \textbf{259.2} & 33.7 & 
54.0 & 57.0 & 88.3 & 53.3 & 82.5 & 67.0 \\

\midrule
UniAD-Base~\cite{hu2023planning} & NC &

45.8 & 16.4 & 129.2 & 43.6 & 
8.9 & 9.3 & 20.0 & 20.0 & 14.2 & 14.5 \\

VAD~\cite{jiang2023vad} & NC &

42.4 & 15.0 & 157.9 & 46.0 & 
11.4 & 11.4 & 18.6 & 20.0 & 19.2 & 18.1 \\

%GenAD~\cite{zheng2024genad} &
%NC &
%44.81 & 15.90 & - & - & - &
%- & - & - & - & - & - \\

MomAD~\cite{song2025don} & NC &

44.5 & 16.7 & 170.2 & \textbf{48.6} & 
- & - & - & - & - & - \\

DriveTransformer-L~\cite{jia2025drivetransformer} & NC &

63.5 & 35.0 & 100.6 & 20.8 & 
17.6 & 35.0 & 48.4 & 40.0 & 52.1 & 38.6 \\

HiP-AD~\cite{tang2025hip}$\dagger$ & NC &

86.8 & 69.1 & 203.1 & 19.4 & 
50.0 & 84.4 & 83.3 & 40.0 & 50.5 & 65.9 \\

Qwen2.5~\cite{Qwen2.5-VL} & NC &

63.9 & 31.6 & 119.3 & 10.1 &
14.3 & 28.9 & 30.1 & 30.0 & 24.7 & 25.6 \\

ORION~\cite{fu2025orion} & NC &

77.7 & 54.6 & 151.5 & 17.4 & 
25.0 & 71.1 & 78.3 & 30.0 & 69.2 & 54.7 \\

ReCogDrive~\cite{li2026recogdrive} & NC &

71.4 & 45.5 & 138.2 & 17.5 & 
29.7 & 20.0 & 69.1 & 20.0 & 71.3 & 42.0 \\

% DiffRefiner~\cite{yin2026diffrefiner} & NC &

% 87.1 & 71.4 & - & - & 
% \textbf{63.8} & 60.0 & 85.0 & 50.0 & \textbf{86.3} & 69.0 \\

GeRo~\cite{yasarla2026generative} & NC &

81.9 & 60.1 & 176.5 & 40.2 & 
40.1 & 78.2 & 87.3 & 50.0 & 76.8 & 66.5 \\
\midrule
\ours (ours)& NC &
85.2
 & 67.5 & 184.3 & 37.9 &  46.3
 & 80.7 & 88.1 & 60.0 & 78.1 & 70.6 \\
\ours$\dagger$ (ours) & NC &
 \textbf{88.3}
 & \textbf{70.5} & 210.3 & 39.0 &  49.4 
 & \textbf{86.2} & \textbf{90.0} & \textbf{60.0} & 80.4 & \textbf{73.2} \\
\bottomrule
\end{tabular}
}
\vspace{-15pt}
\end{table*}

\section{Results}
\vspace{-7pt}
We perform closed-loop evaluation and multi-ability test of \ours on Bench2Drive and Fail2Drive, compare with recent state-of-the-art methods, and evaluate the latent rollout realism of \ours. We further conduct ablation studies on different aspects of our proposed approach. More results can be found in the appendix, including inference computation analysis (Section~\ref{app:inference_analysis}), open-loop evaluation (Section~\ref{app:open-loop}), and additional qualitative visualization results and comparison (Section~\ref{app:vis}).

\vspace{-4pt}
\subsection{Closed-Loop Evaluation}
\vspace{-4pt}
\paragraph{Quantitative Results.}
Table~\ref{tab:b2d_unified} summarizes quantitative comparisons with state-of-the-art methods on Bench2Drive, including closed-loop evaluation over 220 test routes and multi-ability assessments across five challenging urban driving behaviors: merging, overtaking, emergency braking, giving way, and traffic sign handling.
Our method outperforms all existing approaches and achieves state-of-the-art performance across key metrics. Notably, the diverse multi-agent driving behaviors enabled by \ours lead to consistent improvements on most multi-ability metrics, highlighting its robustness and effectiveness in complex interactive driving scenarios. We also report open-loop evaluation results in Appendix Section~\ref{app:open-loop}.

Standard \ours uses temporal waypoint supervision at 0.5\,s intervals and the Bench2Drive controller. \ours$\dagger$ adds geometric path waypoints at 2\,m spacing alongside temporal waypoints and uses the SimLingo PID controller during closed-loop evaluation. This variant aligns waypoint supervision and controller choices with SimLingo and HiP-AD while using the original Bench2Drive 950 training videos. Note that SimLingo trains on a different, larger set of driving scenarios (approximately 650K samples) and a stronger PDM-Lite expert; these training-data differences remain.

\begin{figure}[t]
    \centering
    \includegraphics[width=\linewidth]{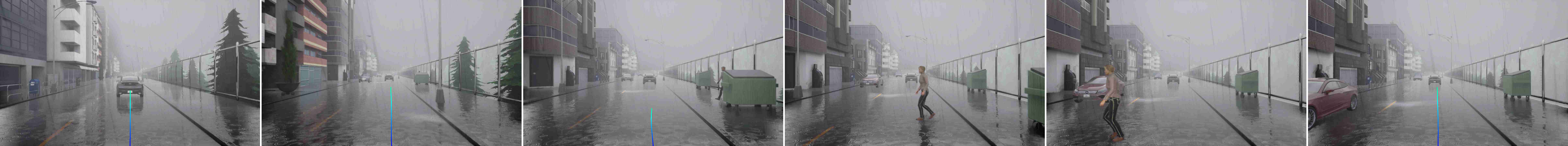}
    \vspace{15pt}
    \includegraphics[width=\linewidth]{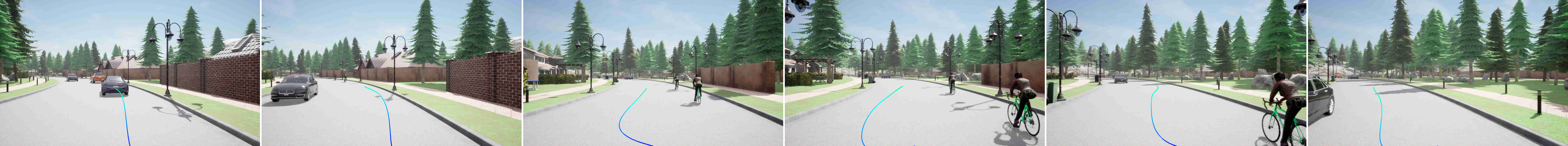}
    \vspace{-30pt}
    \caption{\small \textbf{Qualitative examples of closed-loop driving on Bench2Drive using \ours.} 
    We show representative trajectories in diverse scenarios, including adverse-weather scenes with limited visibility and sudden pedestrian crossings (top row), and clear suburban traffic with dynamic agents such as cyclists and surrounding vehicles (bottom row). 
    Blue curves denote the planned ego-vehicle trajectory, highlighting smooth and adaptive behavior during interactive maneuvers.}
    \vspace{-15pt}
    \label{fig:qualitative}
\end{figure}

% \vspace{-8pt}
\paragraph{Qualitative Results.}
Figure~\ref{fig:qualitative} illustrates the closed-loop performance of the proposed \ours across diverse environments and interaction scenarios. In adverse weather conditions with reduced visibility (top row), the policy demonstrates cautious and stable behavior by smoothly reducing speed and applying mild braking when encountering pedestrians and roadside obstacles. In clear suburban settings with dynamic agents such as vehicles, pedestrians, and cyclists (bottom row), \ours adapts its planned trajectory to safely yield, overtake, and negotiate interactions through gradual steering and acceleration adjustments. Across all scenarios, the planned ego trajectories remain smooth and consistent, indicating coherent action selection and stable closed-loop control under varying environmental conditions. Additional qualitative results can be found in Appendix Section~\ref{app:vis}.

\begin{table}[t!]
% \color{red}
% \captionsetup{labelfont={color=red},textfont={color=red},justification=raggedright,singlelinecheck=false}
\begin{minipage}[t]{0.49\linewidth}
\centering
\captionof{table}{\small Closed-loop evaluation on Fail2Drive, in terms of Driving Score (DS) and Success Rate (SR, in \%); both are higher the better. $\dagger$ uses geometric path supervision and the SimLingo PID controller.}
\vspace{1pt}
\label{tab:fail2drive}
\small
\setlength{\tabcolsep}{4pt}
\resizebox{0.86\linewidth}{!}{%
\begin{tabular}{lcccc}
\toprule
\multirow{2}{*}{Method} & \multicolumn{2}{c}{In-distribution} & \multicolumn{2}{c}{Generalization} \\
\cmidrule(lr){2-3} \cmidrule(lr){4-5}
& DS$\uparrow$ & SR$\uparrow$ & DS$\uparrow$ & SR$\uparrow$ \\
\midrule
ORION & 53.0 & 52.0 & 51.2 & 46.0 \\
SimLingo & 80.6 & 79.3 & 71.7 & 55.0 \\
HiP-AD & 74.1 & 70.7 & 67.1 & 56.7 \\
\ours & 85.3 & 82.1 & 76.1 & 63.7 \\
\ours$\dagger$ & 86.7 & 82.9 & 79.5 & 65.8 \\
\bottomrule
\end{tabular}}
\end{minipage}
\hfill
\begin{minipage}[t]{0.49\linewidth}
\centering
\captionof{table}{\small Realism of latent rollouts on Bench2Drive using adapted Waymo Sim Agents metrics. Higher values indicate better agreement with the reference distribution.}
\label{tab:rollout_realism}
\vspace{1pt}
\small
\setlength{\tabcolsep}{4pt}
\resizebox{0.95\linewidth}{!}{%
\begin{tabular}{lccc}
\toprule
Method & Meta & Kinematic & Interactive \\
\midrule
ORION~\citep{fu2025orion} & 0.391 & 0.263 & 0.518 \\
SimLingo~\citep{renz2025simlingo} & 0.421 & 0.368 & 0.573 \\
DeepSight~\citep{zhang2026deepsight} & 0.490 & 0.413 & 0.608 \\
% Author-computed GeRo scores: meta=0.553499139555786; kinematic=0.454424717; interactive=0.6198703516.
GeRo~\citep{yasarla2026generative} & 0.553 & 0.454 & 0.620 \\
\ours & 0.572 & 0.476 & 0.703 \\
\bottomrule
\end{tabular}}
\end{minipage}
\vspace{-10pt}
\end{table}

\paragraph{Generalization on Fail2Drive}
\label{sec:fail2drive}
Fail2Drive~\citep{Gerstenecker2026Fail2Drive} tests closed-loop generalization using paired CARLA routes: each route containing a distribution shift is matched with an in-distribution counterpart. Its 17 new scenario classes cover appearance, layout, behavioral, and robustness shifts. Reporting both splits therefore examines adaptation to changed driving conditions beyond the standard Bench2Drive scenarios.
Table~\ref{tab:fail2drive} shows that \ours maintains 85.3 DS on the in-distribution split, close to its 85.2 DS on Bench2Drive (Table~\ref{tab:b2d_unified}). On the generalization split, despite trained on less data, \ours reaches 76.1 DS and 63.7\% SR, outperforming SimLingo by 4.4 DS and 8.7 SR points and HiP-AD by 9.0 DS and 7.0 SR points. \ours$\dagger$ further improves these scores to 79.5/65.8. Nevertheless, \ours's SR falls from 82.1\% to 63.7\% across the two splits, showing that substantial generalization challenges remain.

\paragraph{Latent Rollout Realism}
\label{sec:rollout_realism}
Closed-loop driving scores measure the deployed policy's performance, but do not directly assess the fidelity of the latent futures used during training. We therefore evaluate decoded rollout trajectories using metrics adapted from the Waymo Open Sim Agents Challenge~\citep{montali2023simagents}, with ground-truth Bench2Drive trajectories as the reference distribution. All methods are evaluated with the same protocol using their best checkpoint; DeepSight~\citep{zhang2026deepsight} provides a latent world-modeling comparison.

The kinematic metric averages likelihoods of linear speed, linear acceleration, angular speed, and angular acceleration, assessing agreement with reference motion statistics. The interactive metric averages likelihoods of distance to the nearest object, collision indication, and time-to-collision, assessing proximity and collision-risk statistics. The realism meta score linearly aggregates these measures. We omit the map component because distance-to-center values are unavailable for some methods. Thus, the evaluation measures trajectory-distribution agreement rather than latent-token reconstruction or exact matching of a single future.

Table~\ref{tab:rollout_realism} shows gains in both kinematic and interactive agreement, with \ours reaching 0.476 and 0.703, respectively, and a meta score of 0.572. The interactive improvement is particularly relevant because surrounding-agent behavior shapes the safety and progress feedback used in rollout RL. These results complement the driving evaluation by checking the generated trajectories from the latent model.  
% They do not guarantee that every latent future obeys exact physical dynamics or exclude exploitation of imperfect rewards. 
Appendix~\ref{app:rollout_realism} gives further evaluation protocol details and discussion.

\subsection{Ablation Study}
\vspace{-4pt}
\paragraph{Ego-Agent Rollout Setting.}
We perform an ablation study on Bench2Drive closed-loop evaluation to analyze the impact of ego-agent rollout at different training stages as shown in Table~\ref{tab:b2d_ego_agent}. The baseline model does not incorporate future state estimation or rollout. Introducing single-step future state estimation during pretraining improves Driving Score (DS) by 2.3 points over the baseline. Enabling multi-step ego-agent rollout yields further gains, outperforming single-step estimation by 4.2 points in DS.
Incorporating rollout during supervised fine-tuning (SFT) leads to additional improvements, and the best performance is achieved when rollout is applied consistently across pretraining, SFT, and reinforcement learning (RL) stages. Compared to the baseline, this full rollout strategy improves DS by 12.3 points and increases the success rate by 7.9%.
\vspace{-8pt}
\paragraph{Single- vs. Multi-Planner}
As shown in Table~\ref{tab:b2d_diverse_planner}, we ablate on single‑planner, multi‑planner, and multi‑planner with diverse action modeling under different rollout settings. 
Under the ego‑only setting, rollout is applied only to the ego agent, while performance is further improved by adopting a multi‑planner strategy. During inference, the final planner is selected based on the reward likelihood learned during the reinforcement learning stage. The self-play setting applies rollout to both the ego and reactive agents  using a single shared planner $P^e$.
Multi-agent play extends this design by assigning different planners to different reactive agents, drawing from a pool $\{P_k\}_{k=1}^{N_R}$. 
The results consistently show that multi‑planner approaches achieve higher Driving Scores than single‑planner baselines. 
Moreover, multi-planner configurations with diverse action modeling outperform those that share a single planner across agents, and enabling reactive-agent rollout further improves performance compared to ego-only training. Most importantly, the multi-planner setup with diverse action modeling achieves the best overall results, improving Driving Score by an additional 4.8 points and Success Rate by 7.7 points.
\vspace{-8pt}
\paragraph{Additional Results.}
We have conducted more ablation studies to analyze the impact of several design choices, including the number of reactive agents, reward components, and supervised loss terms in SFT. which can be found in Appendix Section~\ref{app:ablation}.
% Diverse action modeling further extends this design by employing multiple planners for different agents.
% The results consistently show that multi‑planner approaches achieve higher Driving Scores than single‑planner baselines. Moreover, multi‑planner configurations using $\{P_k\}_{k=1}^{N_R}$ the pool of action planners used for reactive agents with diverse action modeling outperform those using shared action modeling, and incorporating reactive agent rollout further improves performance compared to ego‑only training. 
% Most importantly, the multi‑planner with diverse action modeling achieves the best overall results, improving Driving Score by an additional 4.5 points and Success Rate by 4.6 points.

\begin{table}[t]
\centering

\begin{minipage}[t]{0.48\linewidth}
\centering
\captionof{table}{\small 
Closed-loop ablation study of ego-only rollout at different training stages on Bench2Drive. 
\emph{Roll} indicates whether ego-agent future rollout is enabled during training. 
SFT and RL denote supervised fine-tuning and reinforcement learning.
}
\label{tab:b2d_ego_agent}

\setlength{\tabcolsep}{6pt}
\renewcommand{\arraystretch}{1.15}

\resizebox{\linewidth}{!}{%
\begin{tabular}{l c c c c c}
\toprule
\textbf{Method} 
& Ego 
& $\mathcal{A}^{\text{react}}$
& Roll
& \textbf{DS} $\uparrow$ 
& \textbf{SR (\%)} $\uparrow$ \\
\midrule
\textit{Baseline}     
& \cmark & \xmark & \xmark & 68.1 & 51.9 \\

\midrule
\textit{+ state estimation pretraining}        
& \cmark & \xmark & \xmark & 70.4 & 54.3 \\

%\textit{+ state estimation rollout}        
%& \cmark & \xmark & \cmark & 72.3 & 55.7 \\

%\midrule
\textit{+ rollout SFT}   
& \cmark & \xmark & \cmark & 77.5 & 57.4 \\

\textit{+ rollout RL for ego-only}   
& \cmark & \xmark & \cmark & \textbf{80.4} & \textbf{59.8} \\ 
\bottomrule
\end{tabular}
}
\end{minipage}
\hfill
\begin{minipage}[t]{0.48\linewidth}
\centering
\captionof{table}{\small 
Closed-loop ablation study of rollout-based action planning strategies on Bench2Drive. Single: all agents share one planner. Multi: agents are assigned distinct planners.
}
\label{tab:b2d_diverse_planner}

\setlength{\tabcolsep}{6pt}
\renewcommand{\arraystretch}{1.15}

\resizebox{\linewidth}{!}{%
\begin{tabular}{l c c c c c}
\toprule
\textbf{Rollout Type} &
\textbf{Planner} &
\textbf{\shortstack{Diversity\\Reg.}} &
\textbf{\shortstack{Reactive\\Agents}} &
\textbf{DS} $\uparrow$ &
\textbf{SR (\%)} $\uparrow$ \\
\midrule
\multirow{3}{*}{Ego Only}
& Single & \xmark & \xmark & 80.4 & 59.8 \\
& Multi  & \xmark & \xmark & 81.8 & 61.5 \\
& Multi  & \cmark & \xmark & 82.6 & 62.7 \\

\midrule
\multirow{2}{*}{Self-Play}
& Single & \xmark & \cmark & 82.1 & 62.0 \\
& Single & \cmark & \cmark & 84.1 & 63.9 \\

\midrule
\multirow{2}{*}{\shortstack{Multi-Agent Play}}
& Multi & \xmark & \cmark & 83.4 & 66.9 \\
& Multi & \cmark & \cmark & \textbf{85.2} & \textbf{67.5} \\
\bottomrule
\end{tabular}
}
\end{minipage}
\vspace{-15pt}
\end{table}
\vspace{-8pt}
% \section{Impact Statement and Limitations}
\section{Conclusion}\label{sec:conclusion}
\vspace{-8pt}
We introduced \ours, a simulator‑free multi‑agent training framework that enables closed‑loop, interactive planning through latent‑space rollouts in vision–language–action models. By extending rollouts to reactive agents and incorporating diversity‑aware reinforcement learning, \ours generates richer interactions and improves robustness in data‑scarce settings. Extensive experiments on Bench2Drive demonstrate state‑of‑the‑art closed-loop performance, highlighting \ours as a scalable and effective approach for advancing end‑to‑end autonomous driving.
\vspace{-8pt}
\paragraph{Limitations.}
\label{sec:limitations}
Despite its strong performance, \ours has a few limitations. First, multi‑agent interactions are modeled through latent rollouts rather than physical simulation. This enables fast and scalable training but can potentially have limited fidelity in scenarios that requires very high‑precision dynamics. 
 % or game‑theoretic agent interactions. 
Second, our experiments were primarily conducted on Bench2Drive. As part of future work, we will extend our experiments to real-world data when robust, accurate real-data-based closed-loop simulators become available.
\vspace{-8pt}
\paragraph{Broader Impact.}
\label{sec:impact_statement}
\ours provides a simulator‑free, multi‑agent closed‑loop training framework that improves the robustness of end-to-end autonomous driving models. Our work can facilitate the development of autonomous driving systems that have improved safety, adaptability, and performance in complex driving scenarios, particularly in data‑scarce or long‑tail situations.

{
    \small
    \bibliographystyle{ieeenat_fullname}
    \bibliography{main}
}

%%%%%%%%%%%%%%%%%%%%%%%%%%%%%%%%%%%%%%%%%%%%%%%%%%%%%%%%%%%%

\appendix

\clearpage

\section{Ablation Study}\label{app:ablation}

\subsection{Number of Reactive Agents}

\paragraph{Agent Distribution in Bench2Drive.}
To contextualize our choice of reactive agent count, we analyze the distribution of traffic agents in the Bench2Drive validation set across 12,806 frames. As reported in Table~\ref{tab:agent_stats}, each frame contains on average 11.91 vehicles (std: 9.59, median: 8, max: 41), 0.06 pedestrians, and 0.03 cyclists.
Vehicles account for over 99\% of all annotated agents. The high variance and gap between mean and median indicate that scene density varies significantly across routes, with some frames containing up to 41 vehicles. These statistics motivate our investigation into the number of reactive agents and confirm that
8 reactive agents provides sufficient coverage for the large majority of evaluation scenarios.

\begin{table}[h!]
\centering
\caption{Per-frame agent statistics in the Bench2Drive validation set (12,806 frames).}
\label{tab:agent_stats}
\setlength{\tabcolsep}{6pt}
\renewcommand{\arraystretch}{1.15}

\resizebox{0.5\linewidth}{!}{%
\begin{tabular}{l c c c c c}
\toprule
\textbf{Category} & \textbf{Mean} & \textbf{Std} & \textbf{Min} & \textbf{Max} & \textbf{Median} \\
\midrule
Vehicles     & 11.91 & 9.59 & 0 & 41 & 8.0 \\
Pedestrians  &  0.06 & 0.33 & 0 &  3 & 0.0 \\
Cyclists     &  0.03 & 0.18 & 0 &  1 & 0.0 \\
\midrule
Total agents & 12.00 & 9.57 & 0 & 41 & 8.0 \\
\bottomrule
\end{tabular}
}
\end{table}

\paragraph{Ablation on the Number of Reactive Agents.}
Reactive agents are selected as those most relevant to the current driving scenario. We study the effect of the number of reactive agents by varying this number among 1, 2, 4, and 8, with agents selected using a KNN-based strategy. As shown in Table~\ref{tab:b2d_agents_num}, the baseline uses no reactive agents with multi-planner and diversity regularization. Increasing the number of reactive agents consistently improves both Driving Score and Success Rate, yielding gains of up to +2.6 DS and +4.8 SR when using 8 reactive agents. Since 8 reactive agents already capture most influential interactions even in dense driving scenarios, further increasing this number is likely to introduce additional computational overhead with limited benefit; therefore, we select 8 reactive agents in all experiments.

\begin{table}[h!]
\centering
\caption{
Ablation study on the number of reactive agents under closed-loop evaluation on Bench2Drive (B2D).
All results are obtained using multi-planner with diversity regularization.
All $\Delta$ values denote improvement \emph{w.r.t. the setting with 0 reactive agents}.
}
\label{tab:b2d_agents_num}

\setlength{\tabcolsep}{8pt}
\renewcommand{\arraystretch}{1.15}

\resizebox{0.5\linewidth}{!}{%
\begin{tabular}{c c c c c}
\toprule
\textbf{\# Reactive Agents}
& \textbf{DS} $\uparrow$
& $\boldsymbol{\Delta}$\textbf{DS}
& \textbf{SR (\%)} $\uparrow$
& $\boldsymbol{\Delta}$\textbf{SR} \\
\midrule

0 & 82.6 & -- & 62.7 & -- \\

1 & 82.8 & +0.2 & 63.4 & +0.7 \\

2 & 83.9 & +1.3 & 65.4 & +2.7 \\

4 & 84.5 &  +1.9 & 66.6 & +3.9 \\

8 & \textbf{85.2} & \textbf{+2.6} & \textbf{67.5} & \textbf{+4.8} \\

\bottomrule
\end{tabular}%
}
\end{table}

\subsection{Losses in SFT}
\label{app:sft_losses}

% \paragraph{SFT loss ablation.}
Table~\ref{tab:sft_loss_ablation} analyzes the effect of applying supervised losses to different entities during SFT. Starting from the setting without explicit SFT supervision (ID~1), adding the ego planning loss (ID~2) provides a clear gain in both Driving Score and Success Rate, indicating improved ego control under closed-loop evaluation. Supervising background-agent motion prediction in addition to ego (ID~3) further improves performance, suggesting that more accurate scene evolution during rollouts benefits ego planning. Finally, adding supervision for reactive-agent planning (ID~4) yields the best results, demonstrating that explicitly training interactive agents alongside the ego leads to stronger closed-loop robustness.

\begin{table}[h!]
\centering
\caption{Ablation of supervised losses used during SFT. ``Ego'' denotes applying the planning loss to the ego trajectory, $\textbf{Z}^{\text{react}}$ denotes supervising reactive-agent planning, and $\textbf{Z}^{\text{bg}}$ denotes supervising background-agent motion prediction.}
\label{tab:sft_loss_ablation}

\setlength{\tabcolsep}{6pt}
\renewcommand{\arraystretch}{1.15}
\resizebox{0.4\linewidth}{!}{%
\begin{tabular}{c c c c c c}
\toprule
\textbf{ID} & \textbf{Ego} & $\boldsymbol{\textbf{Z}^{\text{react}}}$ & $\boldsymbol{\textbf{Z}^{\text{bg}}}$ & \textbf{DS} $\uparrow$ & \textbf{SR (\%)} $\uparrow$ \\
\midrule
(1) & \xmark & \xmark & \xmark & 70.4 & 54.3 \\
(2) & \cmark & \xmark & \xmark & 75.2 & 56.1 \\
(3) & \cmark & \xmark & \cmark & 77.5 & 57.4 \\
(4) & \cmark & \cmark & \cmark & 79.6 & 59.1 \\
\bottomrule
\end{tabular}}
\end{table}

\subsection{Rewards in RL}
Table~\ref{tab:reward_ablation} presents the performance improvements of each reward term in our RL fine-tuning objective. Starting from the base setting SFT model (\ie, without additional rewards), adding the global reward $G_t$ improves both Driving Score (DS) and Success Rate (SR), indicating that scene-level safety feedback provides useful training signal. Incorporating the agent-specific reward $r^i_t$ yields a larger gain, improving DS to 83.4 and SR to 66.9, reflecting better progress and interaction-aware behavior at the agent level. Finally, enabling the diversity reward $D_t$ further boosts performance to the best overall results (DS 85.2, SR 67.5), suggesting that diversity regularization complements safety/progress rewards by mitigating mode collapse and encouraging distinct yet effective behaviors during rollout-based training.
\begin{table}[h!]
\centering
\caption{Ablation on reward components. $G_t$ denotes the global reward, $r^i_t$ the agent-specific reward, and $D_t$ the diversity reward.}
\label{tab:reward_ablation}

\setlength{\tabcolsep}{6pt}
\renewcommand{\arraystretch}{1.15}
\resizebox{0.4\linewidth}{!}{%
\begin{tabular}{c c c c c c}
\toprule
\textbf{ID} & $G_t$ & $r^i_t$ & $D_t$ & \textbf{DS} $\uparrow$ & \textbf{SR (\%)} $\uparrow$ \\
\midrule
(5) & \xmark & \xmark & \xmark & 79.6 & 59.1 \\
(6) & \cmark & \xmark & \xmark & 81.1 & 62.2 \\
(7) & \cmark & \cmark & \xmark & 83.4 & 66.9 \\
(8) & \cmark & \cmark & \cmark & \textbf{85.2} & \textbf{67.5} \\
\bottomrule
\end{tabular}}
\end{table}

\section{Diversity Reward}
\label{app:diversity_reward}

To encourage heterogeneous driving styles (e.g., conservative vs.\ assertive merges) and to mitigate mode collapse during RL fine-tuning, we introduce a diversity reward that promotes behavioral separation among a \emph{group} of sampled rollouts. Concretely, we maintain a pool of reactive-agent policies/planners
$\mathcal{P}=\{\pi_\theta^e\}\cup\{\pi_\theta^{a_j}\}_{j=1}^{N_R}$,
where $\pi^e$ is the ego policy/planner and $\{\pi_\theta^{a_j}\}$ are behavior-specialized policies/planners used to generate trajectories for $N_R$ reactive agents. In all experiments, we set $N_R=8$.

\paragraph{Behavior Descriptor.}
We summarize each rolled-out trajectory segment $\tilde{\tau}_{t:t+T}$ using a descriptor $\Gamma(\cdot)$ formed from the NAVSIM driving-quality metrics~\citep{dauner2024navsim}: No at-fault Collisions (NC), Drivable Area Compliance (DAC), Driving Direction Compliance (DDC), Traffic Light Compliance (TLC), Ego Progress (EP), Time to Collision (TTC) within bound, Lane Keeping (LK), History Comfort (HC), and Extended Comfort (EC). These measures describe safety, rule compliance, progress, and comfort.
% \paragraph{Behavior Descriptor.}
% We summarize each rolled-out agent trajectory using a compact behavior descriptor $\Gamma(\cdot)$ computed from the predicted trajectory segment $\tilde{\tau}_{t:t+T}$. The descriptor concatenates normalized scalar metrics capturing safety, rule compliance, and comfort, including:
% (i) mean longitudinal acceleration,
% (ii) mean jerk,
% (iii) minimum time-to-collision (TTC) along the horizon,
% (iv) lane-centering error,
% (v) lane-change timing (if applicable),
% (vi) drivable-area compliance,
% (vii) driving-direction compliance, and
% (viii) traffic-light compliance.
% Each component is normalized (e.g., by a fixed range or running statistics) so that no single metric dominates the $\ell_1$ distance.

\paragraph{Planner Assignment and Rollout Sampling.}
 For each reactive agent
$a_j \in \textbf{Z}^{\text{react}}_t$, we assign a behavior category (e.g., \emph{aggressive}, \emph{cautious}, \emph{rule-compliant}, \emph{high-comfort} from the NAVSIM driving-quality metrics) and compute the diversity reward $D_t$ for agent $a_j$ accordingly.
For example, if agent $a_7$ is categorized as aggressive and agent $a_2$ as safe/comfortable, $\pi_\theta(a_7)$ and ${\pi_\theta(a_2)}$ generate their respective predicted trajectories $\tilde{\tau}^{a_2}$ and $\tilde{\tau}^{a_3}$. $D_t$ for $a_7$ is computed using the $\Gamma(\cdot)$ aggressive descriptor and $D_t$ for $a_2$ is computed using the $\Gamma(\cdot)$ safe/comfortable descriptor.
This yields predicted trajectories for all reactive agents in the rollout.

\paragraph{Diversity Reward.}
 We define the diversity reward as the average pairwise $\ell_1$ distance between behavior descriptors across the different behavior metrics:
\begin{equation}
\label{eq:div_reward}
D_t
=
\frac{1}{N_R}
\sum_{j=1}^{N_R} \sum_{\Delta=1}^T
\Big\|\Gamma^e\big(\tilde{\tau}^e_{t+\Delta}\big)
-
\Gamma^{a_j}\big(\tilde{\tau}^{a_j}_{t+\Delta}\big)
\Big\|_1,
\end{equation}
This reward assigns higher values when different planner selections induce measurably distinct (yet safe and rule-compliant) behaviors, thereby encouraging the policy to maintain a diverse set of plausible interaction patterns during training.

\section{Inference Computation Analysis}
\label{app:inference_analysis}
% \paragraph{Inference Latency}
Table~\ref{tab:latency} reports the per-frame inference latency of \ours measured on a single NVIDIA H100 (80GB) GPU with batch size 1, using 10 warmup batches followed by 50 timed batches with \texttt{torch.cuda.synchronize()} for accurate GPU timing.
Each timed forward pass encompasses the full inference pipeline: (1) a multi-camera image backbone and neck that extract per-frame visual features from surrounding cameras; (2) a BEV encoder that lifts and fuses perspective features into a bird's-eye-view representation; (3) a Qwen2.5-1.5B language model head that encodes scene context and generates agent-conditioned queries; (4) a map head for local map prediction; and (5) per-agent decoders (comprising present/future distribution encoders, GRU-based trajectory predictors, and ego future decoders) instantiated independently for each of the 8 reactive planners. 
Rollout is disabled at inference, so no environment simulation or autoregressive steps are performed.
With 8 reactive-agent planners, \ours requires 12{,}988 GFLOPs per inference and achieves a mean latency of 487.7\,ms (2.05 FPS), with a P90 of 529.1\,ms.
The compute cost is dominated by the Qwen2.5-1.5B language model, which accounts for over 99\% of total FLOPs, while the image backbone contributes only 15 GFLOPs.
% , demonstrating consistent throughput suitable for offline evaluation and closed-loop simulation.

\begin{table}[h]
\centering
\caption{Computational cost and per-frame inference latency of \ours on a single NVIDIA H100 GPU (batch size 1). Rollout is disabled at inference.}
\label{tab:latency}
\setlength{\tabcolsep}{6pt}
\renewcommand{\arraystretch}{1.15}
\resizebox{0.8\linewidth}{!}{%
\begin{tabular}{l c c c c c c}
\toprule
\textbf{Method} & \textbf{\#Planners} & \textbf{GFLOPs} & \textbf{Mean (ms)} & \textbf{P50 (ms)} & \textbf{P90 (ms)} & \textbf{Inferences Per Second} \\
\midrule
\ours & 8 & 12{,}988 & 487.7 & 491.6 & 529.1 & 2.05 \\
\bottomrule
\end{tabular}
}
\end{table}

\section{Open-Loop Evaluation on Bench2Drive}\label{app:open-loop}
In Table~\ref{tab:b2d_unified_open}, we present open-loop assessment on Bench2Drive, we additionally report the average L2 error between predicted and ground-truth trajectories over a 2‑second horizon at 2Hz.

\begin{table*}[h!]
\centering
\caption{Unified comparison of closed-loop, open-loop, and multi-ability performance on the Bench2Drive base set. Avg.~L2 denotes the average trajectory error over 2 seconds at 2~Hz. NC = Navigation Command, TP = Target Point. $\dagger$ indicates models trained with geometric path waypoint supervision and evaluated using the PID controller from SimLingo.}
\label{tab:b2d_unified_open}
\vspace{-5pt}
\resizebox{0.99\textwidth}{!}{
\begin{tabular}{lcccccccccccccc}
\toprule
Method &
Condition &
\multicolumn{4}{c}{Closed-loop Metric} &
Open-loop &
\multicolumn{6}{c}{Multi-Ability Test(\%) $\uparrow$} \\
\cmidrule(lr){3-6}
\cmidrule(lr){8-13}
 & &
DS$\uparrow$ &
SR(\%)$\uparrow$ &
Efficiency$\uparrow$ &
Comfort$\uparrow$ &
Avg.~L2$\downarrow$ &
Merging &
Overtaking &
E-Brake &
Give Way &
T.Sign &
Mean \\

\midrule
TCP-traj~\cite{wu2022trajectory} &
TP &
59.9 & 30.0 & 76.5 & 18.1 & 1.70 &
28.8 & 24.3 & 51.7 & 40.0 & 46.3 & 34.2 \\

TCP-traj w/o distillation~\cite{wu2022trajectory} &
TP &
49.3 & 20.5 & 78.8 & 22.9 & 1.96 &
28.7 & 28.7 & 48.3 & 40.0 & 28.7 & 34.1 \\

ThinkTwice~\cite{jia2023thinktwice} &
TP &
62.4 & 31.2 & 69.3 & 16.2 & 0.95 &
27.4 & 18.4 & 35.8 & 50.0 & 54.2 & 37.2 \\

DriveAdapter~\cite{jia2023driveadapter} &
TP &
64.2 & 33.1 & 70.2 & 16.0 & 1.01 &
28.4 & 28.4 & 47.5 & 50.0 & 56.4 & 42.1 \\
SimLingo~\cite{renz2025simlingo}$\dagger$ &
TP &
85.1 & 67.3 & 259.2 & 33.7 & - &
54.0 & 57.0 & 88.3 & 53.3 & 82.5 & 67.0 \\

\midrule
UniAD-Base~\cite{hu2023planning} &
NC &
45.8 & 16.4 & 129.2 & 43.6 & 0.73 &
8.9 & 9.3 & 20.0 & 20.0 & 14.2 & 14.5 \\

VAD~\cite{jiang2023vad} &
NC &
42.4 & 15.0 & 157.9 & 46.0 & 0.91 &
11.4 & 11.4 & 18.6 & 20.0 & 19.2 & 18.1 \\

%GenAD~\cite{zheng2024genad} &
%NC &
%44.81 & 15.90 & - & - & - &
%- & - & - & - & - & - \\

MomAD~\cite{song2025don} &
NC &
44.5 & 16.7 & 170.2 & 48.6 & 0.87 &
- & - & - & - & - & - \\

DriveTransformer-Large~\cite{jia2025drivetransformer} &
NC &
63.5 & 35.0 & 100.6 & 20.8 & 0.62 &
17.6 & 35.0 & 48.4 & 40.0 & 52.1 & 38.6 \\

HiP-AD~\cite{tang2025hip}$\dagger$ &
NC &
86.8 & 69.1 & 203.1 & 19.4 & 0.69 &
50.0 & 84.4 & 83.3 & 40.0 & 50.5 & 65.9 \\

Qwen2.5~\cite{Qwen2.5-VL} &
NC &
63.9 & 31.6 & 119.3 & 10.1 & 0.87 &
14.3 & 28.9 & 30.1 & 30.0 & 24.7 & 25.6 \\

ORION~\cite{fu2025orion} &
NC &
77.7 & 54.6 & 151.5 & 17.4 & 0.68 &
25.0 & 71.1 & 78.3 & 30.0 & 69.2 & 54.7 \\

ReCogDrive~\cite{li2026recogdrive} &
NC &
71.4 & 45.5 & 138.2 & 17.5 & - &
29.7 & 20.0 & 69.1 & 20.0 & 71.3 & 42.0 \\
% DiffRefiner~\cite{yin2026diffrefiner} &
% NC &
% 87.1 & 71.4 & - & - & - &
% 63.8 & 60.0 & 85.0 & 50.0 & 86.3 & 69.00 \\
GeRo~\cite{yasarla2026generative} &
NC &
81.9 & 60.1 & 176.5 & 40.2 & 0.57 &
40.1 & 78.2 & 87.3 & 50.0 & 76.8 & 66.5 \\
\midrule
\ours (ours)& 
NC & 85.2
 & 67.5 & 184.3 & 37.9 & 0.54 & 46.3
 & 80.7 & 88.1 & 60.0 & 78.1 & 70.6 \\
\ours (ours)$\dagger$ &
NC & 88.3
 & 70.5 & 210.3 & 39.0 & 0.53 & 49.4 
 & 86.2 & 90.0 & 60.0 & 80.4 & 73.2 \\
\bottomrule
\end{tabular}
}
\vspace{-0pt}
\end{table*}

\section{Rollout Realism and Broader Discussion}
\subsection{Rollout Realism}
\label{app:rollout_realism}
We evaluate predicted trajectories using metrics adapted from the Waymo Open Sim Agents Challenge~\citep{montali2023simagents}, with ground-truth Bench2Drive trajectories as the reference distribution. The kinematic metric averages likelihoods of linear speed, linear acceleration, angular speed, and angular acceleration. The interactive metric averages likelihoods of distance to the nearest object, collision indication, and time-to-collision. The realism meta metric is a linear aggregation of these metrics. We omit the map metric because distance-to-center values are unavailable for some methods. All compared checkpoints are evaluated under the same protocol using the best checkpoint from each method.

Table~\ref{tab:rollout_realism} in the main paper reports the resulting scores. \ours improves all three metrics, including the realism meta score from 0.391 for ORION to 0.572. DeepSight provides a relevant latent world-modeling comparison. These metrics directly assess trajectory-level agreement with reference motion and interaction statistics; they do not guarantee exact physical dynamics for every generated future.

\subsection{Latent Exploration, Model Scale, and Pretrained Knowledge}
\label{app:rollout_discussion}
Imperfect rewards can favor unrealistic latent futures. \ours mitigates this through auxiliary state-estimation pretraining and rollout SFT before RL, together with collision, TTC, progress, and behavior-based rewards. Diversity is measured through driving-quality descriptors rather than arbitrary directions in latent space. The realism evaluation above provides complementary evidence about the resulting trajectories.

Larger VLM backbones may offer stronger semantic priors for rare scenes, but would also increase inference cost. Their effect on rollout realism and long-tail performance has not been evaluated here. The reported inference cost in Appendix~\ref{app:inference_analysis} characterizes the current backbone.

The curriculum improves the controlled ego-only ablation from 68.1/51.9 DS/SR to 70.4/54.3 after state-estimation pretraining, 77.5/57.4 after rollout SFT, and 80.4/59.8 after rollout RL (see Table~\ref{tab:b2d_ego_agent}). These task-specific improvements and the Fail2Drive results indicate useful driving adaptation, but do not directly establish that general pretrained Qwen knowledge is preserved. A dedicated evaluation of forgetting remains future work.

\section{Qualitative Results}\label{app:vis}

\subsection{Qualitative Comparison with ReCogDrive}
\label{subsec:qualitative_recogdrive}

\begin{figure*}[t]
    \centering
    \includegraphics[width=0.7\textwidth]{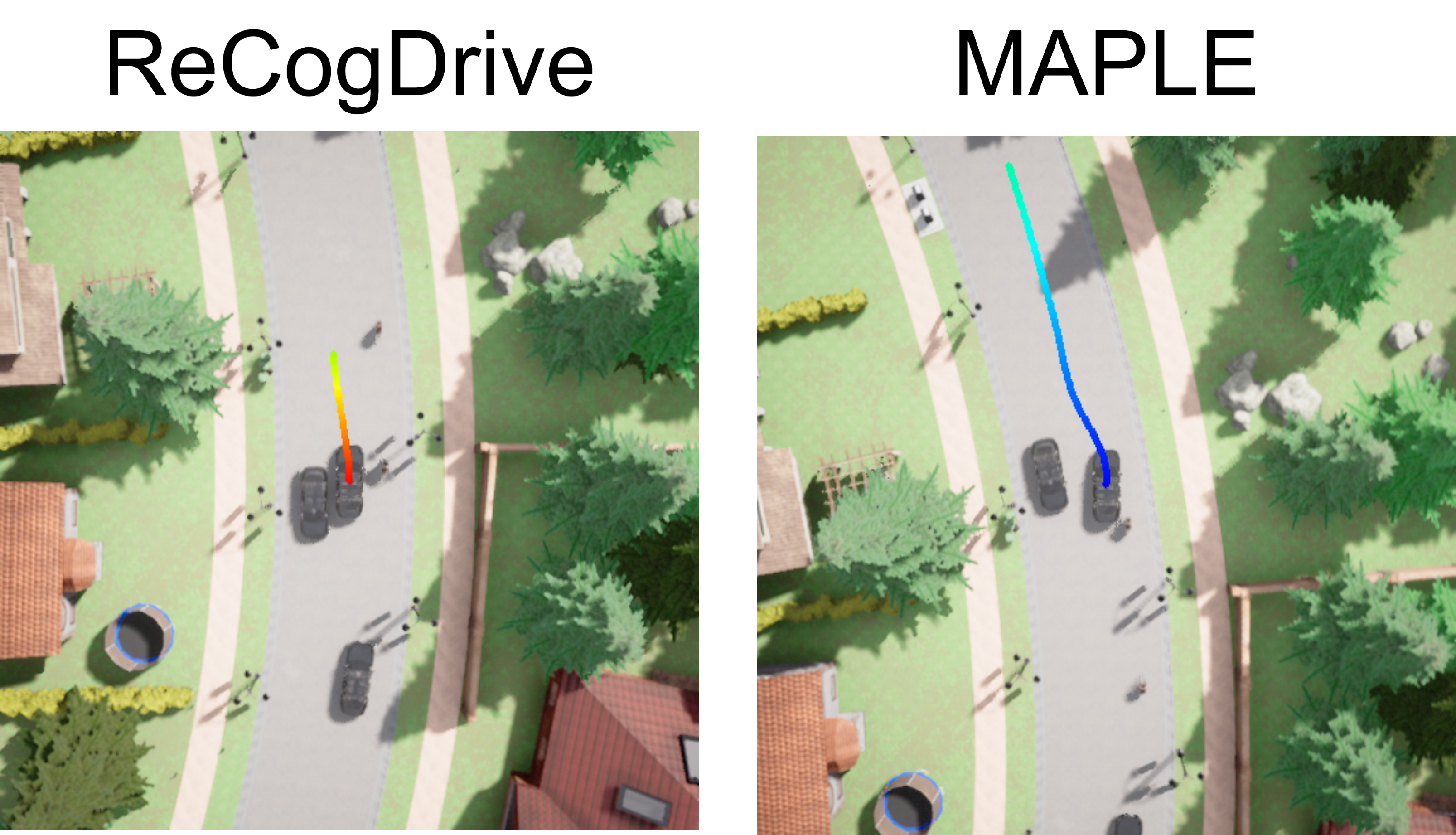}
    \caption{\textbf{BEV qualitative comparison on Bench2Drive (closed-loop).}
    Bird’s-eye-view visualization for the same route/scenario (\texttt{RouteScenario\_25951\_rep0}, \texttt{HazardAtSideLaneTwoWays\_1}, \texttt{weather\_id}=7).
    Left: ReCogDrive~\cite{li2026recogdrive}. Right: \ours (ours). The planned ego trajectory is overlaid, illustrating different interaction outcomes in the same context.}
    \label{fig:bev_recogdrive_vs_maple}
\end{figure*}

\begin{figure*}[t!]
    \centering
    \includegraphics[width=\textwidth]{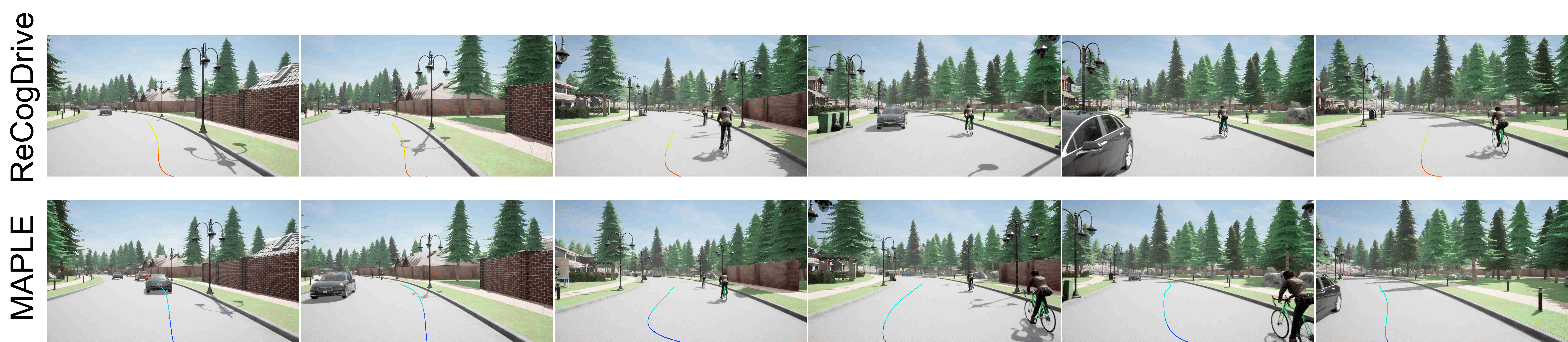}
    \caption{\textbf{Closed-loop rollout comparison on Bench2Drive.}
    Multi-frame qualitative rollouts for the same route/scenario (\texttt{RouteScenario\_25951\_rep0}, \texttt{HazardAtSideLaneTwoWays\_1}, \texttt{weather\_id}=7).
    Top row: ReCogDrive. Bottom row: \ours.
    Colored curves denote the planned ego trajectory across time, highlighting differences in closed-loop interaction behavior.}
    \label{fig:recogdrive_vs_maple}
\end{figure*}

Figures~\ref{fig:bev_recogdrive_vs_maple} and~\ref{fig:recogdrive_vs_maple} qualitatively compare \ours with ReCogDrive~\cite{li2026recogdrive}, which is a very recent work that leverages ego-only, single-step RL to train E2E VLA planners, under closed-loop evaluation on Bench2Drive for the same route. Both methods make progress through an interactive scene containing dynamic agents (e.g., two cyclists traveling along the road) and nearby traffic; however, their interaction outcomes differ. ReCogDrive exhibits a failure case around the 5th column of Figure~\ref{fig:recogdrive_vs_maple}: the planned trajectory deviates abruptly and results in a collision/infraction event, which is also reflected in the BEV visualization in Figure~\ref{fig:bev_recogdrive_vs_maple}. This substantially reduces the composed score (score\_penalty $=0.4225$, score\_composed $=42.25$), despite the route being marked as completed. In contrast, \ours completes the same route without collision infractions (Figures~\ref{fig:bev_recogdrive_vs_maple} and~\ref{fig:recogdrive_vs_maple}) and achieves a perfect composed score (score\_penalty $=1.0$, score\_composed $=100$). Visually, \ours produces smoother and more consistent planned ego trajectories (blue/cyan) with gradual curvature changes while maintaining safe separation from the cyclists and surrounding vehicles, whereas ReCogDrive shows sharper trajectory changes preceding the infraction. Overall, this example highlights that \ours yields more robust closed-loop behavior in interactive scenarios, avoiding safety-critical failures that can occur even when route completion is achieved.


\subsection{Additional Qualitative Results}
\label{app:additional_qual_results}

\begin{figure*}[t]
    \centering
    \includegraphics[width=\textwidth]{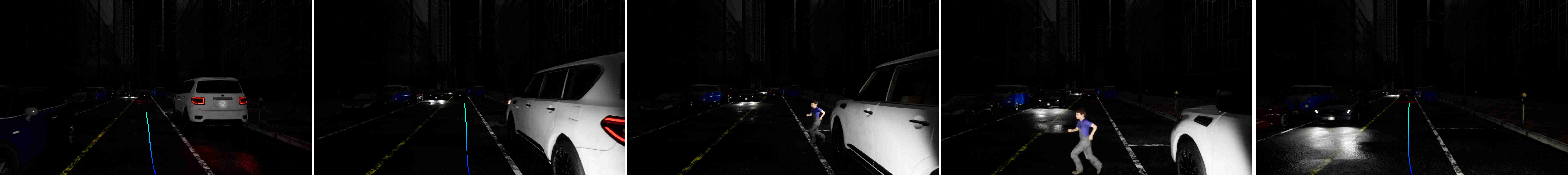}\\[-1mm]
    \includegraphics[width=\textwidth]{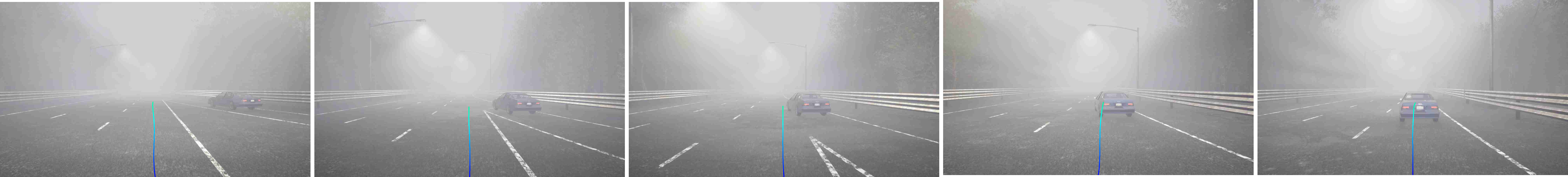}\\[-1mm]
    \includegraphics[width=\textwidth]{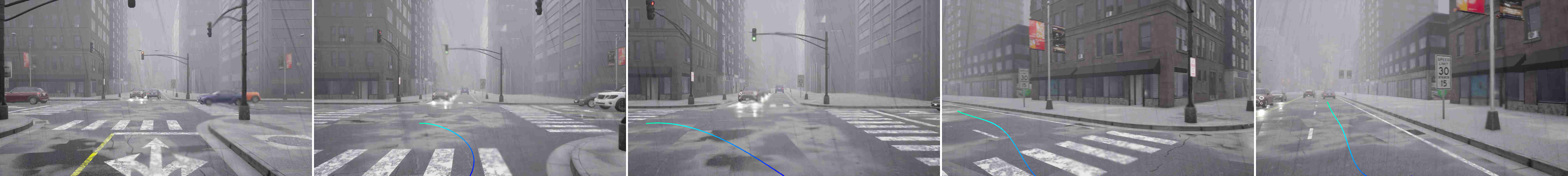}
    \caption{\textbf{Additional qualitative closed-loop driving examples on Bench2Drive using \ours.}
    These examples includes challenging conditions, like low-light/night driving with
    sudden pedestrian appearances and wet-road reflections, dense fog/highway driving with reduced visibility,
    and urban scenes with adverse weather. Blue/cyan curves denote the planned ego trajectory.}
    \label{fig:app_qual_1}
\end{figure*}

\begin{figure*}[t]
    \centering
    \includegraphics[width=\textwidth]{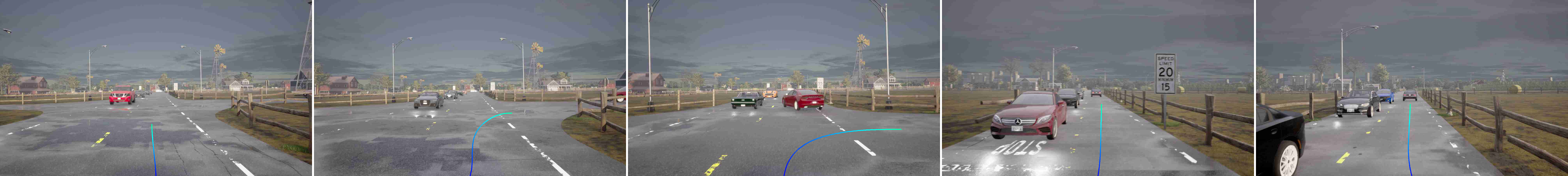}\\[-1mm]
    \includegraphics[width=\textwidth]{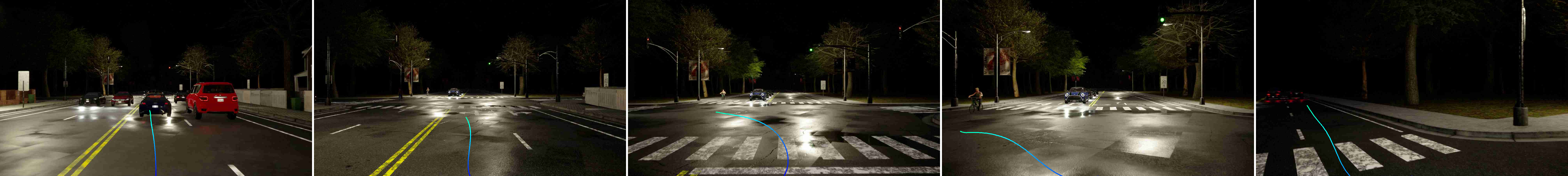}
    \caption{\textbf{Additional qualitative closed-loop driving examples on Bench2Drive using \ours.}
    More examples covering suburban/rural traffic with oncoming vehicles and lane curvature, as well as
    nighttime intersection scenarios with wet-road conditions and surrounding traffic. Blue/cyan curves denote
    the planned ego trajectory.}
    \label{fig:app_qual_2}
\end{figure*}

Figures~\ref{fig:app_qual_1} and \ref{fig:app_qual_2} provide additional qualitative closed-loop driving results on the Bench2Drive benchmark using \ours, covering a broader range of weather, lighting, and interaction patterns. For each example, we show multiple time steps from the ego view with the planned ego trajectory overlaid (blue/cyan). Figure~\ref{fig:app_qual_1} highlights particularly challenging perception and interaction regimes: (i) low-light/night driving on wet roads, where the ego plan remains centered and stable even when a pedestrian suddenly enters near the roadway and nearby vehicles occupy adjacent lanes; (ii) dense fog on a multi-lane road, which substantially reduces visibility and shortens the effective planning horizon, where \ours maintains a smooth, lane-consistent trajectory; and (iii) adverse-weather urban driving at intersections with complex lane markings and cross traffic, where \ours executes a left turn with gradual curvature and without abrupt corrections. Figure~\ref{fig:app_qual_2} further includes suburban/rural scenes with oncoming traffic, vehicles moving at varying speeds, and gentle-to-moderate road curvature, as well as nighttime intersection scenarios under wet-road illumination. Across these diverse settings, \ours consistently produces temporally coherent plans with smooth curvature changes and stable lane positioning, while adapting the trajectory to dynamic agents and roadway structure.


\subsection{Failure Case}
\begin{figure}[t]
    \centering
    \includegraphics[width=\linewidth]{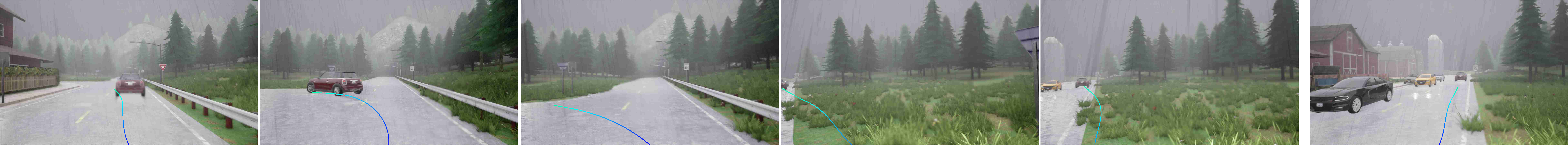}
    \caption{\textbf{Failure case: over-cautious avoidance leading to lane departure (Bench2Drive, closed-loop).}
    In this scenario, \ours performs an overly conservative unprotected left turn to avoid a potential collision, resulting in a brief deviation outside the route lanes for \emph{about 1.0 meters} (\emph{1.29\%} of the full route). The car quickly moves back to the lane after this brief deviation.  Blue/cyan curves denote the planned ego trajectory over time.}
    \label{fig:app_failure_outside_lanes}
\end{figure}

Figure~\ref{fig:app_failure_outside_lanes} shows a failure case where \ours becomes overly cautious during a unprotected left turn and steers toward the road boundary to avoid a potential collision. Resulting in a brief deviation outside the route lanes forabout 1.0\,meters (1.29\% of the completed route). 

\section{Losses}
\label{app:appx_losses}

This section details the loss functions used in the pretraining and supervised fine-tuning (SFT) stages of \ours. We first pretrain the VLA backbone with auxiliary perception tasks to obtain stable scene representations, and then optimize planning and prediction objectives with rollout-based SFT.

\subsection{Pretraining}
\label{app:losses_pretrain}

During pretraining, the VLA model is trained with auxiliary tasks including 3D object detection, map segmentation, agent motion prediction, traffic-light/traffic-state prediction, and future state prediction. These losses encourage the backbone to encode geometry, semantics, and traffic context into latent tokens, stabilizing subsequent rollout training. 

\paragraph{3D Object Detection.}
For detection (via the proposed query-based detector), we use a standard classification + regression objective:
\begin{equation}
\mathcal{L}_{\text{det}} = \mathcal{L}_{\text{cls}} + \mathcal{L}_{\text{reg}},
\end{equation}
where $\mathcal{L}_{\text{cls}}$ is a focal loss and $\mathcal{L}_{\text{reg}}$ is an $\ell_1$ regression loss. 

\paragraph{Traffic-State Prediction.}
For traffic-state (e.g., traffic light / traffic status) prediction, we apply a focal loss:
\begin{equation}
\mathcal{L}_{\text{tra}} = \mathcal{L}_{\text{focal}}.
\end{equation}

\paragraph{Agent Motion Prediction.}
For multi-agent motion prediction, we use a classification + regression loss:
\begin{equation}
\mathcal{L}_{\text{mot}} = \mathcal{L}_{\text{mcls}} + \mathcal{L}_{\text{mreg}},
\end{equation}
where $\mathcal{L}_{\text{mcls}}$ is a focal loss over motion modes and $\mathcal{L}_{\text{mreg}}$ is an $\ell_1$ regression loss on the predicted trajectories. 

\paragraph{Map Segmentation / Vectorized Map Learning.}
For map learning, we use a classification loss (focal loss) and a regression loss based on Manhattan ($\ell_1$) distance between predicted map points and ground-truth map points:
\begin{equation}
\mathcal{L}_{\text{map}} = \mathcal{L}_{\text{map-cls}} + \lambda_{\text{map}} \mathcal{L}_{\text{map-reg}},
\qquad
\mathcal{L}_{\text{map-reg}} = \|\hat{\mathbf{m}}-\mathbf{m}\|_1.
\end{equation}
This follows the practice for vectorized map prediction used in VAD~\cite{jiang2023vad}.

\paragraph{Future State Prediction.}
We additionally supervise future state prediction using the loss in Eq.~\ref{eq:fse_loss}, denoted as $\mathcal{L}_{\text{state}}$.

\paragraph{Overall Pretraining Objective.}
The full pretraining loss is the weighted sum of all components:
\begin{equation}
\label{eq:pretrain_loss}
\mathcal{L}_{\text{pre}}
=
\lambda_{\text{det}}\mathcal{L}_{\text{det}}
+
\lambda_{\text{tra}}\mathcal{L}_{\text{tra}}
+
\lambda_{\text{mot}}\mathcal{L}_{\text{mot}}
+
\lambda_{\text{state}}\mathcal{L}_{\text{state}}
+
\lambda_{\text{map}}\mathcal{L}_{\text{map}}.
\end{equation}

We use the same loss-weight coefficients as VAD~\cite{jiang2023vad} for the auxiliary-task objectives, i.e.,
$\lambda_{\text{det}}$, $\lambda_{\text{mot}}$, and $\lambda_{\text{map}}$ follow the weighting scheme in~\cite{jiang2023vad}. We set $\lambda_{\text{tra}}$, and $\lambda_{\text{state}}$ as 1.0.

\subsection{Planning Loss (SFT / Action Planner)}
\label{app:losses_planner}

The action planner in \ours is trained with a generative objective and trajectory-level supervision. In particular, the planner uses a VAE-style latent alignment loss and trajectory regression, together with safety/feasibility regularizers.

\paragraph{VAE Latent Regularization.}
We use a Kullback--Leibler divergence term to align the latent ``reasoning'' space with the action/trajectory distribution:
\begin{equation}
\mathcal{L}_{\text{vae}} = D_{\mathrm{KL}}\!\left(q(\mathbf{z}\mid \cdot)\,\|\,p(\mathbf{z})\right).
\end{equation}

\paragraph{Trajectory Regression.}
We supervise the predicted waypoints with an MSE (or $\ell_1$) regression loss:
\begin{equation}
\mathcal{L}_{\text{mse}} = \frac{1}{T}\sum_{\Delta=1}^{T}\left\|\hat{\tau}_{t+\Delta}-\tau_{t+\Delta}\right\|_2^2.
\end{equation}

\paragraph{Safety/Feasibility Regularizers.}
Following VAD~\cite{jiang2023vad}, vectorized constraint-based training commonly used in end-to-end driving, we include collision and boundary regularizers to discourage unsafe trajectories:
\begin{equation}
\mathcal{L}_{\text{col}} \;\;\text{(collision avoidance)}, \qquad
\mathcal{L}_{\text{bd}} \;\;\text{(boundary / drivable-area constraint)}.
\end{equation}
These terms penalize trajectories that come too close to other agents or violate road boundaries, consistent with constraint-based planning losses in prior work. 

\paragraph{Overall Planning Loss.}
The final planning objective is:
\begin{equation}
\label{eq:plan_loss}
\mathcal{L}_{\text{plan}}
=
\lambda_{\text{vae}}\mathcal{L}_{\text{vae}}
+
\lambda_{\text{mse}}\mathcal{L}_{\text{mse}}
+
\lambda_{\text{col}}\mathcal{L}_{\text{col}}
+
\lambda_{\text{bd}}\mathcal{L}_{\text{bd}}.
\end{equation}
We use the same loss-weight coefficients as VAD~\cite{jiang2023vad} and ORION~\cite{fu2025orion} for the weights
$\lambda_{\text{vae}}$, $\lambda_{\text{mse}}$, and $\lambda_{\text{col}}$, and $\lambda_{\text{bd}}$.
% \clearpage

% \section{Impact Statement }
% \label{sec:impact_statement}
% This work proposes \ours, a simulator‑free, multi‑agent closed‑loop training framework that improves the robustness and interaction realism of end‑to‑end autonomous driving systems. By enabling scalable multi‑agent play and diverse behavior modeling directly in the latent space, \ours has the potential to improve safety, adaptability, and performance in complex driving scenarios, particularly in data‑scarce or long‑tail situations. 
% At the same time, as with any autonomous driving research, improper deployment or overreliance on learned policies could pose safety risks if models fail under distribution shifts or are applied without adequate system‑level safeguards; therefore, \ours is intended as a research contribution rather than a deployment‑ready solution.

  \section{Training Details}
  \label{app:training_details}

  \subsection{Bench2Drive Training}
The standard Bench2Drive curriculum follows Section~\ref{sec:exp_setup}: 18 epochs of auxiliary pretraining, 24 epochs of rollout SFT, and 6 epochs of RL on 8 NVIDIA H100 GPUs. AdamW uses a learning rate of $2\times10^{-4}$ and weight decay 0.01. We use four input frames, eight reactive-agent planners, and $T=8$ rollout steps with the temporal rates listed in the main paper.

\subsection{Additional nuScenes Fine-Tuning}
\label{app:nusc_finetuning}
The optimizer configuration below applies to an additional nuScenes fine-tuning stage after the standard Bench2Drive curriculum; it is not the Bench2Drive training recipe. We use AdamW~\citep{loshchilov2019decoupled} with $\beta_1=0.9$, $\beta_2=0.999$, and weight decay $10^{-5}$. The base learning rate is $10^{-5}$, with layer-wise decay 0.9, 500 linear warm-up iterations, and a cosine schedule with minimum learning-rate ratio $10^{-3}$. Fine-tuning runs for three epochs on 8 NVIDIA H100 GPUs, with one sample per GPU (effective batch size 8). The reported rollout configuration is eight steps at 2\,Hz.

  \subsection{Reactive Agent Selection.}
  Reactive agents are the traffic participants whose future trajectories are jointly predicted with the ego during self-play rollout.
  At each rollout step, agents are ranked by their distance to the ego vehicle and the $N_R$ nearest neighbors are designated as reactive; the remaining agents are treated as background agents motion trajectories predicted using motion prediction head.

  \subsection{Rollout Configuration.}
  Rollout is performed over $N_{\text{step}}{=}8$ steps at 2\,Hz (a 4-second horizon).
  At each step, the ego trajectory and reactive-agent trajectories are predicted by the model and used to advance the scene state; the resulting state is fed back as input for the next step.
  During inference, rollout is disabled and a single forward pass is executed per frame.

  \subsection{Multi-Planner Setup.}
  We instantiate $N_R$ behavior-specialized planners drawn from a shared pool $\mathcal{P}=\{P^e\}\cup\{P^{a_j}\}_{j=1}^{N_R}$.
  In all reported experiments we use $N_R{=}8$ reactive-agent planners (see Appendix~\ref{app:ablation} for ablations).
  During RL fine-tuning, the best-performing planner for the ego is selected based on the reward likelihood learned in the GRPO stage.
\section{Open-loop evaluation on nuScenes}
Table~\ref{tab:nusc_finetuning} reports open-loop planning under the PARA-Drive evaluation protocol~\citep{weng2024drive}. Zero-shot \ours uses the Bench2Drive-trained checkpoint; the fine-tuned model includes the additional nuScenes stage. Fine-tuning reduces L2 trajectory error from 0.30 to 0.22 and collision rate from 0.14 to 0.12.

\begin{table}[h!]
% \color{red}
% \captionsetup{labelfont={color=red},textfont={color=red}}
\centering
\caption{nuScenes open-loop planning under the PARA-Drive\cite{weng2024drive} evaluation protocol. Lower is better. Zero-shot \ours is trained on Bench2Drive; \ours additionally fine-tunes on nuScenes.}
\label{tab:nusc_finetuning}
\small
\begin{tabular}{lcc}
\toprule
Method & L2 trajectory error$\downarrow$ & Collision rate$\downarrow$ \\
\midrule
UniAD~\citep{hu2023planning} & 0.95 & 0.40 \\
VAD~\citep{jiang2023vad} & 0.91 & 0.30 \\
PARA-Drive~\citep{weng2024drive} & 0.66 & 0.17 \\
SparseDrive~\citep{sun2024sparsedrive} & 0.65 & 0.14 \\
ORION~\citep{fu2025orion} & 0.69 & 0.29 \\
\ours (zero-shot) & 0.30 & 0.14 \\
\ours & 0.22 & 0.12 \\
\bottomrule
\end{tabular}
\end{table}

%%%%%%%%%%%%%%%%%%%%%%%%%%%%%%%%%%%%%%%%%%%%%%%%%%%%%%%%%%%%

\end{document}